\documentclass[lettersize,journal]{IEEEtran}
\usepackage{amsmath,amsfonts}
\usepackage{algorithm}
\usepackage{array}
\usepackage[caption=false,font=normalsize,labelfont=sf,textfont=sf]{subfig}
\usepackage{textcomp}
\usepackage{stfloats}
\usepackage{url}
\usepackage{verbatim}
\usepackage{graphicx}
\usepackage{cite}
\usepackage{multirow}
\usepackage{bbm}
\usepackage{algpseudocode}
\usepackage{color}
\newcommand{\equ}{Eq.}
\newcommand{\tab}{Tab.}
\newcommand{\etal}{et al.}
\newcommand{\alg}{Alg.}
\definecolor{mygray}{gray}{0.5}

\begin{document}

\title{Strong-Weak Integrated Semi-supervision for Unsupervised Single and Multi Target Domain Adaptation}

\author{Xiaohu Lu and {Hayder Radha,~\IEEEmembership{Fellow,~IEEE} }

}

\markboth{Journal of \LaTeX\ Class Files,~Vol.~14, No.~8, August~2021}%
{Shell \MakeLowercase{\textit{et al.}}: A Sample Article Using IEEEtran.cls for IEEE Journals}


\maketitle

\begin{abstract}
Unsupervised domain adaptation (UDA) focuses on transferring knowledge learned in the labeled source domain to the unlabeled target domain. Despite significant progress that has been achieved in single-target domain adaptation for image classification in recent years, the extension from single-target to multi-target domain adaptation is still a largely unexplored problem area. In general, unsupervised domain adaptation faces a major challenge when attempting to learn reliable information from a single unlabeled target domain. Increasing the number of unlabeled target domains further exacerbate the problem rather significantly. In this paper, we propose a novel \textit{strong-weak integrated semi-supervision (SWISS)} learning strategy for image classification using unsupervised domain adaptation that works well for both single-target and multi-target scenarios. Under the proposed SWISS-UDA framework, a strong representative set with high confidence but low diversity target domain samples and a weak representative set with low confidence but high diversity target domain samples are updated constantly during the training process. Both sets are fused randomly to generate an augmented strong-weak training batch with pseudo-labels to train the network during every iteration. The extension from single-target to multi-target domain adaptation is accomplished by exploring the class-wise distance relationship between domains and replacing the strong representative set with much stronger samples from peer domains via \textit{peer scaffolding}. Moreover, a novel adversarial logit loss is proposed to reduce the intra-class divergence between source and target domains, which is back-propagated adversarially with a gradient reverse layer between the classifier and the rest of the network. Experimental results based on three popular benchmarks, Office-31, Office-Home, and DomainNet, show the effectiveness of the proposed SWISS framework with our method achieving the best performance in both Office-Home and DomainNet with improvement margins of 0.4\% and 1.0\%, respectively.

\end{abstract}

\begin{IEEEkeywords}
Domain adaptation, unsupervised, multi-target, peer scaffolding, intra-class divergence.
\end{IEEEkeywords}

\section{Introduction}
\label{introduction}
\IEEEPARstart{T}{he} success of deep neural networks in tackling critical tasks, such as image classification~\cite{he2016deep}, object detection~\cite{ren2016faster,hnewa2023integrated}, semantic segmentation~\cite{badrinarayanan2017segnet}, and image captioning is highly dependent on the availability of large amount of labeled training samples. Meanwhile, the generalization ability of deep neural networks is poor when applied to different target domains. The inclusion of additional labeled samples for the target domains can be a straightforward solution to this deficiency; but such a solution is arguably inefficient and costly.

Unsupervised domain adaptation (UDA) tackles this problem by transferring the knowledge learned in the labeled source domain to the unlabeled target domain. This transference is usually accomplished by aligning the distributions of data points in source domain and target domain such that the classifier trained on the source domain can also be applied onto the target domain. Most current domain alignment methods can be classified into two categories: moment matching based and adversarial learning based. The idea of moment matching is based on the observation that two distributions are similar if their moments in different orders are all close to each other~\cite{peng2019moment}. The Maximum Mean Discrepancy (MMD)~\cite{gretton2007kernel} approach is widely used by this type of methods, which attempt to align domains through minimizing the distance between weighted sums of all raw moments. Another popular paradigm is leveraging the idea of adversarial learning~\cite{xia2021adaptive} that is rooted in Generative Adversarial Networks (GANs)~\cite{goodfellow2014generative}. This approach is based on (a) training a domain classifier to align domains' distributions; yet (b) trying to trick or confuse the domain classifier in differentiating between source and target domain data by generating domain-invariant features. Hence, the adversarial learning is usually performed between the feature generator and the domain classifier so that when the domain classifier is fully confused by the feature generator the goal of domains' alignment is achieved. Despite the fact that most of recent works on domain adaptation still focus mainly on the the single-target scenario, it is noteworthy that multi-target domain adaptation has been drawing increasing attention.


Multi-target UDA~\cite{yang2020heterogeneous,wang2020attention,gholami2020unsupervised,nguyen2021unsupervised,isobe2021multi,roy2021curriculum} assumes that there exists only one source domain, but the target domain samples are collected from several different distributions. The goal of multi-target UDA is to transfer the knowledge learned from the source domain to diverse target domains. Another objective is to explore the underlying relationships and patterns among target domains, such that the performance of each target domain in multi-target UDA is better than that of training several single-source single-target UDA networks separately. Currently, there exist only few works on multi-target UDA, including the attention guided method~\cite{wang2020attention,yang2020heterogeneous}, mutual information based method~\cite{gholami2020unsupervised}, knowledge distillation based method~\cite{nguyen2021unsupervised}, and collaborative consistency learning~\cite{isobe2021multi}. 

Despite the fact that significant progress in single-target UDA and some noteworthy advancement in multi-target domain adaptation areas have been achieved individually, a unified framework that can work for both areas is still missing. In this paper, we address this problem by proposing a strong-weak integrated semi-supervision (SWISS) learning strategy that maintains a strong representative set with high confidence but low diversity samples and a weak representative set with low confidence but high diversity samples in the full training process. Furthermore, the proposed SWISS framework generates augmented training samples with pseudo-labels from these two representative sets to train the network in each iteration. Both single-target and multi-target UDA can be handled in a viable manner under the proposed framework. The extension from single-target to multi-target UDA is achieved by adopting the \textit{peer scaffolding} strategy~\cite{vygotsky1980mind}, which updates the strong representative set with samples not only from the target domain itself but also from its peer domains. Moreover, we also propose a novel adversarial logit loss to improve the performance by reducing the intra-class divergence between source and target domains. In summary, our contributions in this paper\footnote{This paper represents a significantly expanded version of our recent conference paper~\cite{lu2022strong}. This includes generalizing our SWISS framework to address the problem of multi-target unsupervised domain adaptation. Another new contribution of this paper is the novel UDA approach that is based on scaffolding theory.}:
    \begin{itemize}
      \item Developing a unified framework for both single and multiple target domain adaptation via strong-weak integrated semi-supervision.
      \item Adopting the peer scaffolding theory~\cite{vygotsky1980mind}, which has been developed and studied in the general field of education, to learn from peer domains in the multi-target UDA scenario.      
      \item Introducing a novel adversarial logit loss function that can reduce the intra-class domain divergence.
      \item Demonstrating that the proposed SWISS framework achieves state-of-the-art performance in several public benchmarks.
    \end{itemize}

\section{Related Work}
\label{relatedwork}

\subsection{Unsupervised Domain Adaptation}
Most early unsupervised domain adaptation methods in the area of image classification utilized moment matching strategy to align data distributions in the source and target domains. For example, in the work of~\cite{tzeng2014deep}, they used one kernel function and one adaptation layer for moment matching. Long~\etal~\cite{long2015learning} extended this strategy to include multiple kernels and adaptation layers, which led to better performance. Later on, Ganin~\etal~\cite{ganin2016domain} introduced the idea of adversarial learning to domain adaptation by training a binary domain classifier to distinguish the data points from different domains. Based on this idea, most of the later methods explored additional information to further help domains' alignment. For example, in the work of~\cite{long2018conditional}, the prediction scores were combined with the feature vector of the data as conditional information to align the domains. In the work of~\cite{du2021cross}, the gradient discrepancy between source samples and target samples is minimized to improve the accuracy by introducing semantic information of target samples. Similarly, in another work of~\cite{liang2021domain}, an auxiliary classiﬁer only for target data is designed to improve the quality of pseudo labels, which is then utilized for training the network via cross-entropy loss. Meanwhile, in the work of~\cite{na2021fixbi}, training samples from augmented-domains between the source domain and target domain are generated by a fixed ratio-based mixup strategy, which are applied along with two confidence based learning methodologies to train the model. While some of these methods have demonstrated promising results on public benchmarks, directly applying them to multi-target domain adaptation may not lead to performance improvements \cite{isobe2021multi}. Therefore, it is essential to develop dedicated approaches that specifically address the challenges posed by multi-target domain adaptation.

\subsection{Multi-target Domain Adaptation}
The topic of multi-target unsupervised domain adaptation (UDA) has received relatively less attention, with only a few works~\cite{yang2020heterogeneous,wang2020attention,gholami2020unsupervised,nguyen2021unsupervised,isobe2021multi,roy2021curriculum} addressing this specific problem. Most of these works primarily focus on the association between target domains. For example, in the work of~\cite{yang2020heterogeneous}, a unified subspace common for all domains with a heterogeneous graph attention network is learned to propagate semantic information among domains. Similarly, in the work of~\cite{wang2020attention} an attention guided method is proposed to capture the context dependency information on transferable regions among the source and target domains. While in~\cite{nguyen2021unsupervised}, a multi-teacher knowledge distillation method is proposed to iteratively distill the target domain knowledge to a common space. Moreover, in the work of~\cite{gholami2020unsupervised}, a unified mutual information based approach is developed to learn both shared and private information in different domains. Recently, in the work of~\cite{roy2021curriculum}, the graph convolutional network is adopted to aggregate features across domains, which is cooperating with a curriculum learning strategy to improve the final performance in multi-target domain adaptation. Despite the progress made in this field, there still exists a significant performance gap between these multi-target domain adaptation methods and the state-of-the-art single-target domain adaptation methods. Further research and innovation are needed to bridge this gap and develop more effective approaches for multi-target UDA.

\subsection{Semi-supervised Learning}
Similar to unsupervised domain adaptation, semi-supervised learning~\cite{hu2021simple,cai2021exponential,abuduweili2021adaptive} also focuses on tackling labeled and unlabeled samples. In order to make use of unlabeled data, semi-supervised learning methods assume that there exist some underlying relationships between distributions of data. Based on this assumption, several categories of methods are developed. Pseudo-label based methods~\cite{hu2021simple} select high-confidence predictions as the label for unlabeled samples. Information maximization based methods~\cite{hu2017learning} consider that a good distribution should be individually certain and globally diverse, and utilize the information maximization loss to regularize the unlabeled samples. Regularization and normalization based methods~\cite{abuduweili2021adaptive,cai2021exponential} adopt regularization and normalization strategies, e.g., batch normalization, to reduce the model's bias to the source domain such that the model's performance in target domain can be improved. Recently, an increasing number of researchers in unsupervised domain adaptation seek to borrow ideas from semi-supervised learning. For example, \cite{choi2019pseudo} assigned pseudo-labels with highest confidence to unlabeled samples, while in the work of~\cite{liang2020we} the information maximization loss is adopted to improve performance of domain adaptation.


\subsection{Peer Scaffolding}
The concept of peer scaffolding finds its roots in the field of education, where it has been widely employed to enhance students' learning experiences through support from others. This approach is based on the idea that learners can benefit from interactions with more knowledgeable individuals, such as teachers or peers who excel in a particular domain. Peer scaffolding has been extensively studied and evaluated in educational settings, particularly in collaborative activities within classrooms. Previous research has demonstrated the effectiveness of peer scaffolding in various educational contexts. For example, in the domain of writing, Riazi et al. \cite{riazi2011teacher} found that peer scaffolding activities significantly improved students' writing quality during the revision process. Additionally, the work of~\cite{pifarre2010promoting} suggests that incorporating peer scaffolding systems into pedagogical approaches can enhance students' learning processes. Overall, the underlying principle of peer scaffolding is to leverage the knowledge and support of others to facilitate individual learning and achieve better learning outcomes.

\section{Methodology}
\subsection{Network Architecture}

    \begin{figure}
    	\centering
    	\footnotesize
    	\begin{tabular}{c}
    		\includegraphics[width=0.97\linewidth]{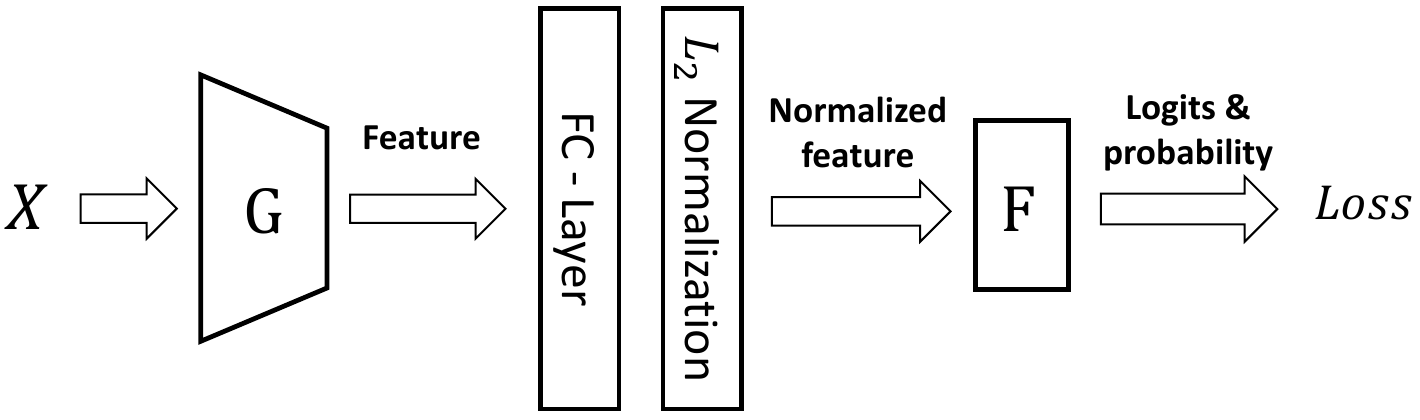}
    	\end{tabular} 
    	\caption{The high-level architecture of the baseline network used in our unsupervised domain adaptation method. For forwarding, training samples are fed into a generator \textbf{G}, followed by a bottleneck network formed by a Fully Connected (FC) layer and a $L_{2}$ normalization layer, and a classifier \textbf{F}.}
    	\label{Fig:basicnet}
    \end{figure}

Fig.~\ref{Fig:basicnet} shows the high-level architecture of our \textit{baseline network} that includes a feature generator network \textbf{G}, a bottleneck network formed by a Fully Connected (FC) layer and a $L_{2}$ normalization layer, and a classifier network \textbf{F}. For \textbf{G}, we employ the pre-trained ResNet-50 or ResNet-101~\cite{he2016deep} as the backbone. For the bottleneck network, the output dimension of the FC-layer is 1024, the $L_{2}$ normalization layer normalizes and scales the feature vector of each input training sample into a normalized feature vector with the $L_{2}$ norm being a constant value $T$. The classifier network $\textbf{F}$ is formed by $k$ sub-classifiers corresponding to the $k$ classes. In the forwarding path, for an image x from the input batch $X$, regardless of its domain, we firstly feed it into \textbf{G} to generate a 1024-d feature vector \textbf{G}(x). Then, we calculate the logits $\{l_1,l_2,...,l_k\}$ between the normalized feature and each sub-classifier using the inner product, and estimate the prediction probabilities of the normalized feature belonging to each class as $\{p_1,p_2,...,p_k\} = \sigma(\{l_1,l_2,...,l_k\})$ with $\sigma$ being the softmax function. After that, we calculate the loss terms based on the logits and probabilities. 

\subsection{Strong-weak Integrated Semi-supervision for Single-target UDA}
\label{single_target}

\begin{figure}
    	\centering
    	\footnotesize
    	\begin{tabular}{c}
    		\includegraphics[width=0.97\linewidth]{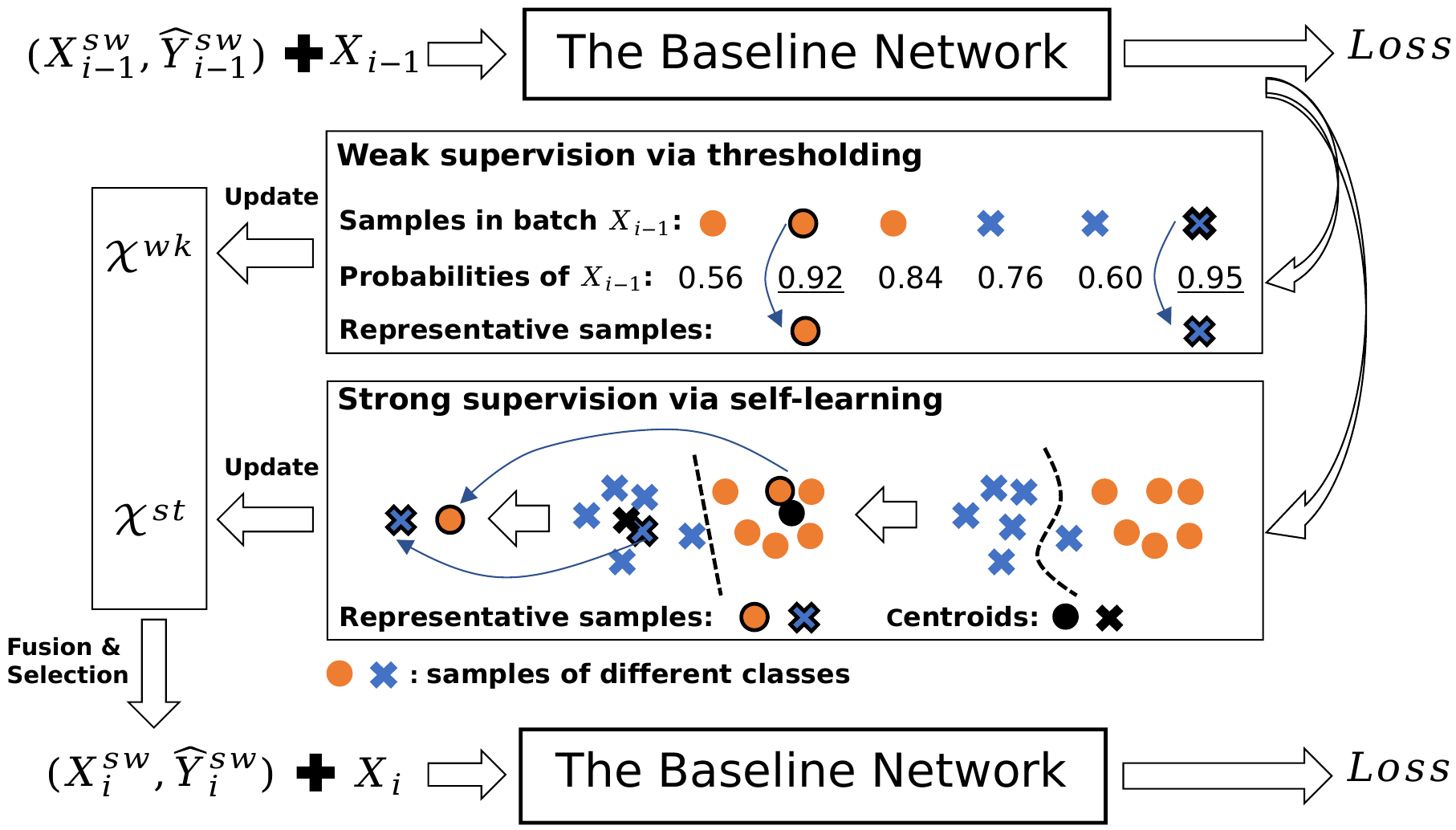}
    	\end{tabular} 
    	\caption{The pipeline of our strong-weak integrated semi-supervision method for typical single-source single-target domain adaptation. A weak representative set $\mathcal{X}^{wk}$ and a strong representative set $\mathcal{X}^{st}$ are maintained during the training process. A strong-weak integrated semi-supervision set of training samples $(X_{i}^{sw}, \hat{Y}_{i}^{sw})$ is generated to train the baseline network.}
    	\label{Fig:framework}
    \end{figure}

Fig.~\ref{Fig:framework} shows the pipeline of our strong-weak integrated semi-supervision method for the typical single-source single-target domain adaptation. The key idea is to obtain a strong representative set $\mathcal{X}^{st}$ and a relatively weak representative set $\mathcal{X}^{wk}$ and then fuse them to generate reliable and diverse augmented samples to train the network. The weak representative set $\mathcal{X}^{wk}$ is updated by target samples with prediction probability higher than a threshold after every iteration. While the strong representative set $\mathcal{X}^{st}$ is formed by target samples with the highest confidence of belonging to each class, which is updated every pre-specified number of iterations via self-learning. In the $i_{th}$ iteration of training, the samples of each class in $\mathcal{X}^{wk}$ and $\mathcal{X}^{st}$ are fused to form an augmented supervision set with pseudo-labels $(X_{i}^{sw}, \hat{Y}_{i}^{sw})$, which along with the target-domain batch $X_{i}$ are utilized to train the baseline network. The rational for this strong-weak integrated semi-supervision is that the strong representative set $\mathcal{X}^{st}$ is of highest prediction confidence but lower diversity, while the weak representative set $\mathcal{X}^{wk}$ is less reliable in prediction but with much higher in diversity. Hence, the combination of both sets gives a good balance between prediction confidence and diversity for the augmented training samples.

\subsubsection{Strong Supervision via Self-learning}
We adopt the same self-learning strategy as in~\cite{du2021cross} to get the strong representative set $\mathcal{X}^{st}$. Namely, given the target domain sample set ${\{x_i\}}_{i=1}^{n}$ with $n$ samples and the baseline network as shown in Fig.~\ref{Fig:basicnet}, there are three steps.
      Firstly, the prediction probabilities and the normalized feature vectors of training samples in the target domain are obtained by feeding the target domain samples into the baseline network. The concatenation of these probabilities $P$ is a $n\times k$ matrix, and that of the normalized feature vectors $V$ is a $n\times d$ matrix. Here $n$ is the number of target domain samples, $k$ is the number of classes, $d=1024$ is the dimension of feature vector.
      
      Then, the initial centroid of class $j$ is obtained as:
            \begin{equation}\label{equ:centroids}
            c_j^{(0)} = \frac{P_{,j}^{T} V}{\sum_{j}{P_{,j}}},
            \end{equation}
      where $P_{,j}$ denotes the $j_{th}$ column of the probability matrix, which corresponds to class $j$. After that, the initial self-supervised pseudo-label can be obtained by assigning each sample to the class with closest centroid as:
            \begin{equation}\label{equ:pseudo_label}
            \hat{y}^{(0)} = arg\min_{j} D(v, c_j^{(0)})
            \end{equation} 
      where $v$ represents the normalized feature vector of a target domain sample, and $D(,)$ measures the cosine distance between two vectors.
      
      Finally, the class centroids are recalculated by replacing the probability matrix $P$ with a $n\times k$ one-hot distribution matrix $\mathbbm{1}$ obtained from the initial self-supervised pseudo-label $\hat{y}^{(0)}$, and the strong representative set $\mathcal{X}^{st}$ is obtained by selecting the target samples closest to each class centroids. Namely, for a specific class $j$,
            \begin{equation}\label{equ:strong_update}
            \begin{split}
            c_j^{(1)} = & \frac{\mathbbm{1}_{,j}^{T} V}{\sum_{j}{\mathbbm{1}_{,j}}}, \\
            x^{st}_j = & x_{arg\min_{i\in[1,n]} D(v_i, c_j^{(1)})}
            \end{split}
            \end{equation}
      where $\mathcal{X}^{st}=\{x^{st}_j\}_{j\in[1,k]}$, $n$ is the number of target samples, $k$ is the number of classes.

Under this self-learning strategy, the distribution of the pseudo-labels will be slightly shifted toward the target domain by assigning each target sample to the closest class' center in a self-supervised way. This self-learning procedure is performed every pre-specified number of iterations, e.g., 200 iterations, and only the samples with closest distances to the centroids of each class are selected as shown in Fig.\ref{Fig:framework}. Note that those strong samples usually have high prediction confidence, which means that they are generally closer to the source domain than other target domain samples. Beside, they are also robust against noise because of the robustness of centroids. For example, some outlier target domain samples may have high prediction probabilities by accident, but they will not be selected as strong representative samples in our method because they are far away from the centroids in the feature space. Based on these two properties, we consider this set of target domain samples as the ``strong" one in comparison with the weak supervision set that is introduced below. 

\subsubsection{Weak Supervision via Thresholding}

For weak supervision, we employ a higher updating frequency than the strong supervision. And, we adopt a simple thresholding strategy to update $\mathcal{X}^{wk}$ for each batch of target samples. More specifically, after inputting a target batch $X=\{x_i\}_{i=1}^{b_s}$ with size $b_s$ to the baseline network, we identify the highest probabilities $\textit{\textbf{p}}=\{p_i\}_{i=1}^{b_s}$ and the corresponding pseudo-labels $\textit{\textbf{l}}=\{l_i\}_{i=1}^{b_s}$ for the $b_s$ samples in the batch. Then, for each class $j$ that has an entry in the pseudo-label vector $\textit{\textbf{l}}$, the sample with the highest probability (which must be greater than a threshold $\lambda$) of belonging to this class $j$ is selected as the new representative sample for that class $j$. In other words, under the weak supervision learning process, there is a single and unique representative sample ${x}^{wk}_j$ for any class $j$ at any given time. Hence, a new sample could potentially be used to update the corresponding representative sample ${x}^{wk}_j$ for class $j$ in the current weak set $\mathcal{X}^{wk}$. Namely,
    \begin{equation}\label{equ:weak_update}
    \begin{split}
    x^{wk}_j &= x_{arg\max_{i} p_i}, \\ \text{s.t.} \  l_i &=j, \  p_i>\lambda,
    \end{split}
    \end{equation}
where $\mathcal{X}^{wk}=\{x^{wk}_j\}_{j\in[1,k]}$, $k$ is the number of classes.
    
\subsubsection{Strong-weak Fusion}
\label{sec:strong_weak_fusion}
In practice, we find that simply using the strong supervision can only slightly reduce the domain divergence. This is because the network can easily over-fit on the strong representative set by remembering them instead of learning knowledge from them due to the small amount of samples in $\mathcal{X}^{st}$ (one sample per class). Meanwhile, in the case of weak supervision only, the discriminability of the network maybe impaired under challenging scenarios due to the error associated with predicting the pseudo-labels in $\mathcal{X}^{wk}$. Fusing both $\mathcal{X}^{st}$ and $\mathcal{X}^{wk}$ can achieve a good balance between them, and consequently, such fusion achieves smaller domain divergence and higher discriminability. Therefore, we adopt a random fusion strategy to generate the fused strong-weak representative set $\mathcal{X}^{sw}$:
    \begin{equation}\label{equ:fusion}
    x^{sw}_j = rx^{st}_j + (1-r)x^{wk}_j, \ \text{s.t.} \ r\in(0, 1),
    \end{equation}
where $x^{st}_j$ and $x^{wk}_j$ are the representative images for class $j$, $x^{sw}_j$ is the fused one in the pixel domain, $r$ is a random number in (0, 1). The reason for using a random value instead of a constant one for $r$ is to increase the diversity of the strong-weak representative set $\mathcal{X}^{sw}$. In practice, we find that adding the full set of $\mathcal{X}^{sw}$ along with the corresponding pseudo-label as supervision may override the target domain itself when the number of classes $k$ is much greater than the batch size, e.g., $k$=345 in DomainNet. Therefore, we only select a small batch $X^{sw}$ from $\mathcal{X}^{sw}$ according to the data distribution of the input batch. Namely, given the input batch $X$ and its most-likely class label $\hat{Y}$ which is obtained from the estimated probability, we have
    \begin{equation}\label{equ:selection_single}
    X^{sw} = \{\mathcal{X}^{sw}_i\}_{i\in \hat{Y}}.
    \end{equation}
Finally, the selected batch of images along with the corresponding label $(X^{sw}, \hat{Y}^{sw})$ are utilized to train the baseline network.

\subsubsection{Pipeline for Single-target UDA}
\label{sec:subsub_pipeline_single}
The pipeline for our strong-weak supervision for single-target domain adaptation can be found in \alg~\ref{alg:single}. Note that at the beginning of training, both $\mathcal{X}^{st}$ and $\mathcal{X}^{wk}$ are empty, which are only utilized for training after being initialized in the first updating of the strong representative set. Details about the loss functions used in \alg~\ref{alg:single} will be introduced in Sec.~\ref{loss_functions}.

    \begin{algorithm}
    \small
    \caption{Strong-Weak Integrated Semi-Supervision (SWISS) for single-target domain adaptation}\label{alg:single}
    \begin{algorithmic}
        \Require a labeled source domain $S$, an unlabeled target domain $T$
        \Ensure optimized parameters of a baseline network as Fig.~\ref{Fig:basicnet}
        \State initialize generator \textbf{G} with pre-trained ResNet
        \For{each iteration}
            \State $L_{CE} \gets$ forward a source batch $(X^{s},Y^{s})$ to network
            \State Back-propagate $L_{CE}$, optimization
            \State $L_{IM}, \ L_{ALL} \gets$ forward a batch $X^{t}$ to network
            \State $(X^{sw}, \hat{Y}^{sw}) \gets$ sample fusion and selection
            \State $L_{SW} \gets$ forward $(X^{sw}, \hat{Y}^{sw})$ to network
            \State Back-propagate $k1L_{IM}$+$k2L_{ALL}$+$k3L_{SW}$
            \State $\mathcal{X}^{wk} \gets$ weak representative set updating
            \If{reach certain iterations}
                \State $\mathcal{X}^{st} \gets$ strong representative set updating 
            \EndIf
        \EndFor
    \end{algorithmic}
    \end{algorithm}

\subsection{Strong-weak Integrated Semi-supervision for Multi-target UDA}
\label{multi_target}
To extend a typical domain adaptation method from single-target to multi-target, a straightforward approach would be to combine all the target domains into a single one. However, it has been noted in prior work \cite{yang2020heterogeneous} that simply combining samples from all target domains can actually degrade performance, as the divergence between certain target domains can be larger than the divergence between the source domain and each target domain individually. To address this challenge, we draw inspiration from the concept of ``peer scaffolding" in the education field \cite{riazi2011teacher}. This involves allowing each target domain to learn not only from the source domain but also from peer domains that share similarities with the target domain. By identifying these peer domains, we can update the strong representative set $\mathcal{X}^{st}$ accordingly, incorporating stronger samples from relevant peer domains. Furthermore, for the weak representative set $\mathcal{X}^{wk}$, we employ the same thresholding strategy as described in Sec.~\ref{single_target}.

    \begin{figure*}
    	\centering
    	\footnotesize
    	\begin{tabular}{c}
    		\includegraphics[width=0.70\linewidth]{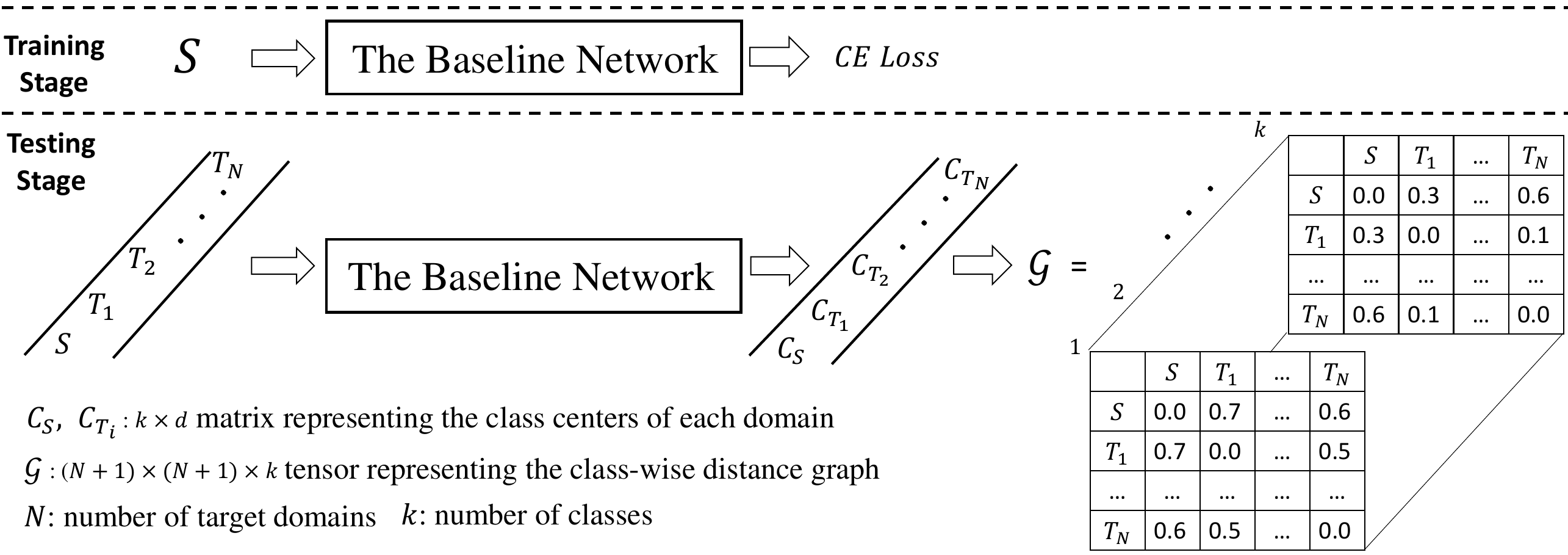}
    	\end{tabular} 
    	\caption{A demonstration on how we apply a two-stage strategy to get the class-wise distance graph for peer scaffolding. A baseline network on the labeled source domain in the training stage. Then each domain is fed into the trained baseline network to calculate the class centers. Finally, a class-wise distance graph is generated to represent the distance between different domains.}
    	\label{Fig:peer_graph}
    \end{figure*} 
    
\subsubsection{Class-wise Peer Scaffolding}
\label{sec:peerscaffolding}
In the context of peer scaffolding~\cite{riazi2011teacher}, a crucial aspect is determining the ``more knowledgeable peers" from whom to learn. In the domain adaptation scenario, the notion of ``more knowledgeable" typically refers to domains that exhibit smaller divergence from the source domain. Some popular domain divergence measures include the KL-divergence~\cite{daume2006domain}, covariance~\cite{sun2016deep}, $\mathcal{H}$-divergence~\cite{ben2010theory}, and Maximum Mean Discrepancy~\cite{gretton2012kernel}. However, most of them are designed to estimate the overall domain divergence, which may not be accurate when the domain divergence shows some class-relevant properties. In this work, we find that the cosine distance between centroids of different classes can be a good measure for \textit{class-wise peer scaffolding}. More specifically, given a labeled source domain $S$, and $N$ unlabeled target domains $\{T_i\}_{i=1}^{N}$, we use the following two-stage strategy to obtain a $(N+1)\times (N+1)\times k$ tensor $\mathcal{G}$ to represent the class-wise distance graph between each pair of domains:
    \begin{itemize}
      \item For the training stage, we train a baseline network on the labeled source domain with the widely used Cross Entropy loss. To avoid over-fitting on the source domain, we stop the training if the prediction accuracy on the source domain starts to decrease.
      \item For the testing stage, we feed each domain's samples into the trained network to get a $n\times k$ probabilities matrix $P$ and a $n\times d$ feature matrix $V$. Here $n$ and $k$ denote the number of samples and classes in the domain, respectively. Then the centroid of each class is obtained in the same way as introduced in \equ~\ref{equ:centroids}. The combination of those $k$ centroids forms a $k\times d$ centroid matrix $C$. Then, given the centroid matrices of source domain $C_S$ and all the target domains $\{C_{T_i}\}_{i=1}^{N}$, a $(N+1)\times (N+1)\times k$ tensor $\mathcal{G}$ can be obtained by calculating the cosine distance between centroid vectors of each pair of domains class-wisely as shown in Fig.~\ref{Fig:peer_graph}. Note that the first row and first column of the $(N+1)\times (N+1)$ symmetric matrix for each class represent the distance of a specific domain to the source domain, while the diagonal elements denote the distance from each domain to itself, which are all zero.
    \end{itemize}

    \begin{figure}
    	\centering
    	\footnotesize
    	\begin{tabular}{c}
    		\includegraphics[width=0.6\linewidth]{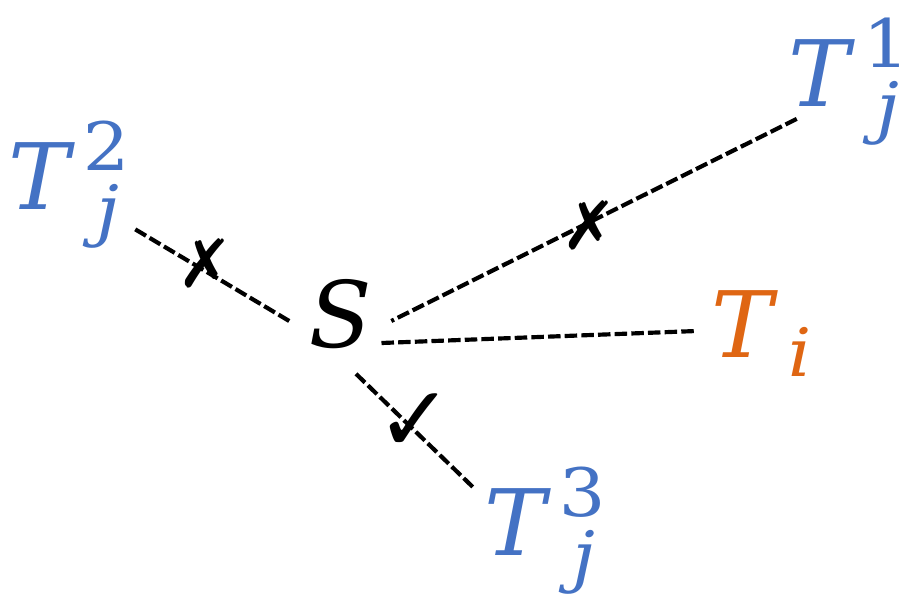}
    	\end{tabular} 
    	\caption{A demonstration on the two criteria for peer scaffolding. Location 3 of $T_j$ satisfies both criteria, therefore it can be used to reduce the domain divergence between $S$ and $T_i$.}
    	\label{Fig:peer_domain}
    \end{figure} 
    
Recall that for single-target domain adaptation, the strong representative set $\mathcal{X}^{st}$ is formed by target samples close to the source domain, which can be easily extended to the multi-target domain case by selecting much closer samples from peer domains based on the class-wise distance tensor $\mathcal{G}$. More specifically, let's consider a given target domain $T_i$. We are trying to identify another target domain, say $T_j$, which might be "more knowledgeable" about the source domain $S$ than $T_i$. In that context, we apply the following two criteria to decide whether the $j_{th}$ target domain is ``more knowledgeable" about the source domain than the $i_{th}$ target domain for a given class $l$: 
    \begin{itemize}
      \item \textit{Criterion 1}: $\mathcal{G}[0, j+1, l] < \mathcal{G}[0, i+1, l]$, which implies that the $j_{th}$ target domain $T_j$ is closer to the source domain than the $i_{th}$ target domain $T_i$ for class $l$.
      \item \textit{Criterion 2}: $\mathcal{G}[i+1, j+1, l] < \mathcal{G}[0, i+1, l]$, which implies that the $j_{th}$ target domain $T_j$ is in-between the source domain and the $i_{th}$ target domain $T_i$ for class $l$.
    \end{itemize}
Fig.~\ref{Fig:peer_domain} gives a demonstration on how these two criteria work for three possible scenarios regarding the distance among target domain $T_i$, target domain $T_j$, and the source domain $S$. The distance to source domain from the first scenario-location of $T_j$ is larger than $T_i$, which violates the first criterion. While for the second scenario location of $T_j$, it's on the other side of the source domain, therefore the second criterion is violated. The third scenario location satisfies both criteria, consequently $T_j$ can be utilized as an in-between domain to help reduce the divergence between $S$ and $T_i$.

Given the class-wise distance tensor $\mathcal{G}$, the updating strategy for strong representative set in the multi-target scenario can be found as follows. For a given class $l\in [1,k]$ in target domain $T_i$, we replace the original strong representative sample in $\mathcal{X}^{st}_i$ with another one from the peer domains which satisfy the two criteria above. More specifically, 
\begin{equation}\label{equ:combined_set}
    \mathcal{X}^{st}_i(l)=\left\{
                \begin{array}{ll}
                  \mathcal{X}^{st}_{j}(l) \ \text{if} \ \text{criterion 1\& 2 satisfied}\\
                  \mathcal{X}^{st}_i(l), \ \text{otherwise} \\
                \end{array}
                \right.
\end{equation}
where $\mathcal{X}^{st}_i(l)$ represents the sample with label $l$ in the strong representative set of the $i_{th}$ target domain. Note that in the case where exists multiple representative samples from peer domains that satisfy the criteria, we randomly choose one from them to replace the corresponding sample in $\mathcal{X}^{st}_i$ in order to improve the diversity of the representative samples. After that, a subset of training samples and their pseudo labels $(X^{sw}, \hat{Y}^{sw})$ are selected out in the same way as \equ~\eqref{equ:selection_single} to train the network. The framework for multi-target UDA is the same as that for single-target UDA as shown in Fig.~\ref{Fig:framework} with the only difference on the updating strategy of the strong supervision set as introduced above.


\subsubsection{Pipeline for Multi-target UDA}
In practice, we find that training $N$ baseline networks for all the $N$ target domains in parallel makes the system complex. Therefore, we train each target domain separately using our single-target pipeline as shown in Fig.~\ref{Fig:framework}, and then select samples with prediction probability greater than a threshold $\lambda$ from each class to form a pseudo strong representative set $\hat{\mathcal{X}}^{st}_i$ for each target domain $T_i$. Consequently, in the fusion and selection procedure, we slightly modify the multi-target fusion strategy in \equ~\eqref{equ:combined_set} by replacing the strong representative set $\mathcal{X}^{st}_{j}$ with the pseudo strong representative set $\hat{\mathcal{X}}^{st}_j$. Then a baseline network is trained on the source domain and applied on the source domain and all the $N$ target domains to get the class-wise distance graph $\mathcal{G}$. After that, for each target domain in the multi-target UDA scenario, a single-target pipeline as shown in Fig.~\ref{Fig:framework} is applied to train a baseline network with the target domain itself, the source domain, and the pseudo strong representative sets from peer domains.

The full pipeline of our strong-weak integrated semi-supervision multi-target domain adaptation is formed by three parts as shown in \alg~\ref{alg:multi}.
    \begin{algorithm}
    \small
    \caption{Strong-Weak Integrated Semi-Supervision for multi-target domain adaptation}\label{alg:multi}
    \begin{algorithmic}
        \Require a labeled source domain $S$, $N$ unlabeled target domains $\{T_i\}_{i=1}^{N}$
        \Ensure $N$ optimized baseline networks as Fig.~\ref{Fig:basicnet}
        \State \textbf{Part 1}
        \For{$i$ $\in$ [1, $N$]}
            \State pseudo $\hat{\mathcal{X}}^{st}_i \gets$ train a single-target UDA network via \alg~\ref{alg:single} 
        \EndFor
        
        \State \textbf{Part 2}
        \State Train a baseline network with $S$ only
        \State $\mathcal{G} \gets$ class-wise distance graph
        
        \State \textbf{Part 3}
        \State Given $\mathcal{G}$, $\{\hat{\mathcal{X}^{st}}_i\}_{i=1}^{N}$
        \For{$i$ $\in$ [1, $N$]}
            \For{each iteration}
                \State $L_{CE} \gets$ forward a source batch $(X^{s},Y^{s})$ to the $i_{th}$ baseline network
                \State Back-propagate $L_{CE}$, optimization
                \State $L_{IM}, \ L_{ALL} \gets$ forward a batch $X^{t}$ to the $i_{th}$ baseline network
                \State $(X^{sw}, \hat{Y}^{sw}) \gets$ sample replacing, fusion, selection
                \State $L_{SW} \gets$ forward $(X^{sw}, \hat{Y}^{sw})$ to the $i_{th}$ baseline network
                \State Back-propagate $k1L_{IM}$+$k2L_{ALL}$+$k3L_{SW}$, optimization
                
                \State $\mathcal{X}^{wk}_i \gets$ weak representative set updating
                \If{reach certain iterations}
                    \State $\mathcal{X}^{st}_i \gets$ strong representative set updating for $T_i$
                \EndIf
            \EndFor
        \EndFor
    \end{algorithmic}
    \end{algorithm}

\subsection{Loss Functions}
\label{loss_functions}

\subsubsection{Source Domain Loss Functions}
For a labeled source batch, we utilize the standard Cross Entropy loss $L_{CE}$ to train the network to minimize the classification error as follows:
    \begin{equation}\label{equ:l_ce}
    L_{CE} = \frac{1}{b_s}\sum_{i=1}^{b_s} \sum_{j=1}^{k} -y_{ij}log(p_{ij})
    \end{equation}
where $b_s$ is batch size, $k$ is the number of classes, $y_{ij}=1$ if the (ground truth) label of the $i_{th}$ sample is $j$, otherwise $y_{ij}=0$.

\subsubsection{Target Domain Loss Functions}
For the unlabeled target domain, we adopt the mutual information maximization loss $L_{IM}$~\cite{li2021transferable,liang2020we} as the baseline and propose a new adversarial logit loss as an improvement. The $L_{IM}$ is based on the mutual information $I(X;\hat{Y})$ between the input $X$ and the output $\hat{Y}$, namely, $I(X;\hat{Y})=H(\hat{Y})-H(\hat{Y}|X)$. $L_{IM}$ is formulated as:
    \begin{equation} \label{equ:l_im}
    \begin{split}
    L_{IM}  = \sum_{j=1}^{k}\hat{p_j}log(\hat{p_j}) + \frac{1}{b_s}\sum_{i=1}^{b_s} \sum_{j=1}^{k}-p_{ij}log(p_{ij}),
    \end{split}
    \end{equation}
where $b_s$ is batch size, $k$ is the number of classes, $p_{ij}$ is the probability of a sample $i$ belonging to class $j$, $\hat{p_j}=\frac{1}{b_s}\sum_{i=1}^{b_s}p_{ij}$ is the average probabilities of samples belonging to class $j$ in a training batch with $b_s$ samples. Note that the mutual information between $X$ and $\hat{Y}$ can also be reformulated as: $I(X;\hat{Y})=H(X)-H(X|\hat{Y})$, which means that minimizing $H(X|\hat{Y})$ can also improve the performance. Given that $H(X|\hat{Y}) = \sum_{j=1}^{k}p_j H(X|\hat{Y}=j)$, we can see that the final goal is to minimize the conditional entropy of $X$ given that it belongs to class $j$. 

Recall that in our baseline network, the feature vector $v$ fed into the classifier $\textbf{F}$ is normalized by a constant temperature $\tau$. Therefore, the equation for the logit $l_j$ between $v$ and the $j_{th}$ prototype can be reformulated as: 
\begin{equation}\label{equ:logit}
l_j = \textbf{w}_j^{\text{T}} \cdot v  = T||\textbf{w}_j^{\text{T}}||cos(\theta),
\end{equation}
where $\textbf{w}_j$ is the $j_{th}$ prototype, which is also the $j_{th}$ weight vector in the classifier $\textbf{F}$, $\theta$ is the angle between $v$ and $\textbf{w}_j$. We can see from the equation above that for a given class $j$, the prototype $\textbf{w}_j$ is constant for all the samples belonging to this class, therefore, a feasible way to reduce the intra-class variance of feature vectors is to minimize $\theta$ for those samples. In order to reduce the angle $\theta$ for target domain samples, we propose a novel adversarial logit loss which tries to enlarge $cos(\theta)$ by reducing $\textbf{w}_j$. More specifically, we formulate the adversarial logit loss $L_{ALL}$ as:
    \begin{equation}\label{equ:l_all}
    L_{ALL}=\frac{1}{b_s} \sum_{i=1}^{b_s} 
        \begin{cases}
            0, \text{otherwise} 
        \end{cases}
    \end{equation}
where $b_s$ is the batch size, $l_{ij}$ is the logit value between sample $i$ and class $j$, $p_{ij}$ is the corresponding prediction probability. \equ~\eqref{equ:l_all} implies that for a given input target domain sample, we firstly calculate its logits to each class in the forwarding path, then select the largest one as its contribution toward the overall value of the proposed adversarial loss $L_{ALL}$. Subsequently, $L_{ALL}$ is back-propagated with a gradient reverse layer between the classifier \textbf{F} and the rest of the network such that \textbf{F} will try to minimize $L_{ALL}$ by reducing the norm of prototypes, while the bottleneck layers and $\textbf{G}$ focus on maximizing $L_{ALL}$ by decreasing the angle $\theta$.

    \begin{figure}
    	\centering
    	\footnotesize
    	\begin{tabular}{c}
    		\includegraphics[width=1.0\linewidth]{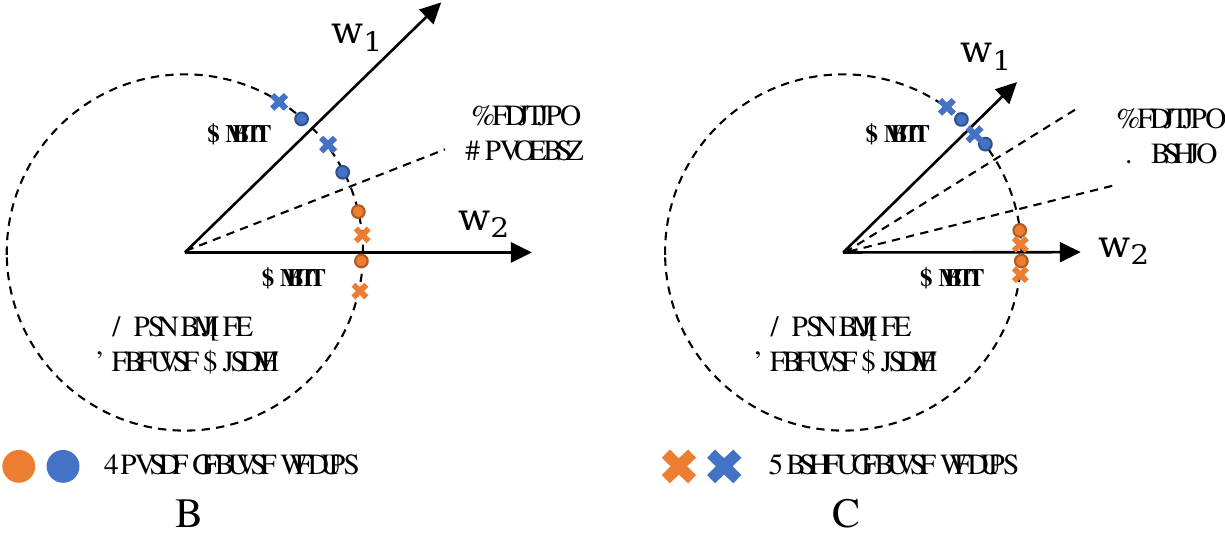}
    	\end{tabular} 
    	\caption{ A 2D illustration of the effect of the proposed adversarial logit loss $L_{ALL}$, (a) before $L_{ALL}$ intra-class variance and domain divergence are large, (b) after $L_{ALL}$ both intra-class variance and domain divergence are reduced. Note that due to the $L_2$ normalization, all the feature vectors are lying on the normalized feature circle with radius being $\tau$ ($\tau$ is the temperature).}
    	\label{Fig:all_loss}
    \end{figure} 

\subsubsection{Strong-weak Integrated Semi-supervision Loss Functions}
    \begin{figure}
    	\centering
    	\footnotesize
    	\begin{tabular}{c}
    		\includegraphics[width=0.80\linewidth]{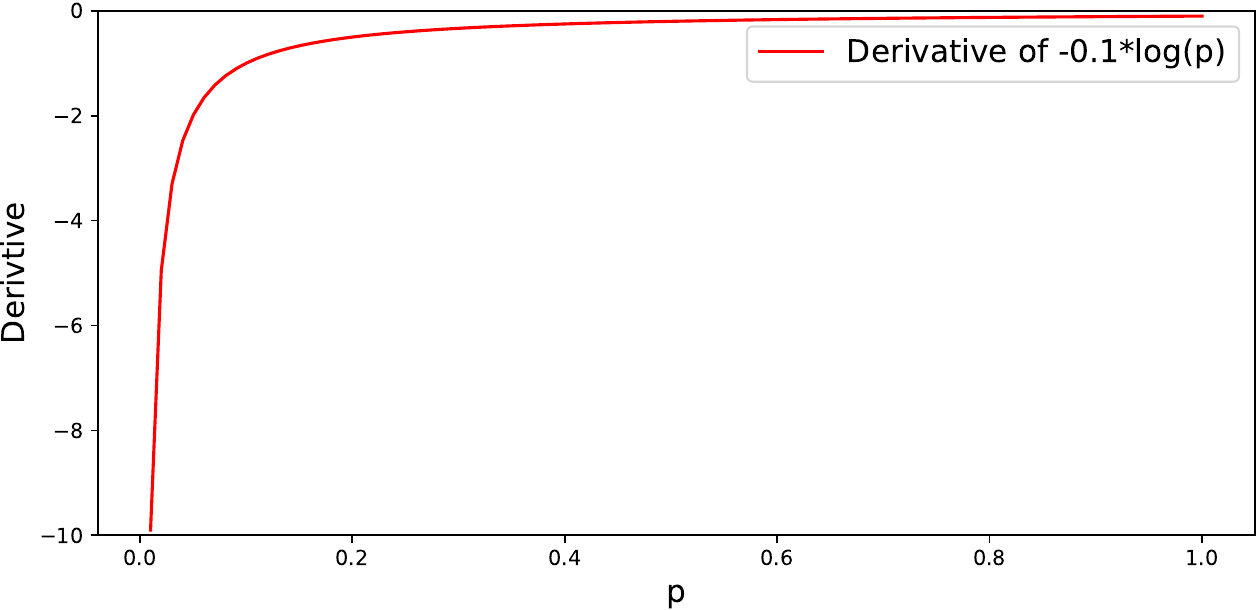}
    	\end{tabular} 
    	\caption{Demonstration on the derivative of $-0.1*log(p)$. The derivative is close to 0 when $p$ is large.}
    	\label{Fig:ce_derivative}
    \end{figure} 
For the strong-weak integrated semi-supervision batch $(X^{sw}, \hat{Y}^{sw})$, one straightforward approach is to use the Cross Entropy as the loss function. However, as shown in Fig.~\ref{Fig:ce_derivative}, we found that the derivative of the Cross Entropy is relatively small when the probability is approaching 1.0, which means that it's hard for the Cross Entropy loss to optimize the network when the pseudo-label is incorrectly estimated with a high confidence. Therefore, we design the following function for the strong-weak integrated semi-supervision loss $L_{SW}$:
    \begin{equation}\label{equ:l_sw}
    L_{SW} = \frac{1}{b_s} \sum_{i=1}^{b_s} \sum_{j=1}^{k} y_{ij} (1.0-p_{ij})
    \end{equation}
which replaces $-log(p_{ij})$ in \equ~\eqref{equ:l_ce} with $1.0-p_{ij}$ such that the derivative is constant regardless of the probability.

As a summary, the overall optimization goal is:
    \begin{equation}\label{equ:final_goal}
    minimizing \begin{cases}
        L_{CE} \ for \ S\\
        k1L_{IM}+k2L_{ALL}+k3L_{SW} \ for \ T.
    \end{cases}
    \end{equation}
Details about the hyper-parameters $k1$, $k2$, $k3$ will be discussed in the experimental section.

\section{Experiments}
 
 \subsection{Setup}

 We term our strong-weak integrated semi-supervision for single-target and multi-target domain adaptation as SWISS(single) and SWISS(multi), respectively. For quantitative assessment of the proposed method, we evaluate it on three popular benchmarks with more than two domains for unsupervised domain adaptation: Office-31~\cite{saenko2010adapting}, Office-Home~\cite{venkateswara2017deep},  DomainNet~\cite{peng2019moment}; and, we compare our proposed method with state-of-the-art single-target and multi-target methods.

\textbf{Office-31}~\cite{saenko2010adapting} is a small-size benchmark that contains 31 classes of objects under the office environment. Three domains: Amazon (A) with 2817 samples, DSRL (D) with 498 samples, and Web (W) with 795 samples are included.

\textbf{Office-Home}~\cite{venkateswara2017deep} is a medium-size benchmark which contains 65 classes of objects under both office and home environments. Four domains: Art (Ar) with 2427 samples, Clipart (Cl) with 4365 samples, Product (Pr) with 4439 samples, and Real-World (Re) with 4357 samples are included. The Office-Home benchmark is much more challenging than the Office benchmark because samples in Art and Product domains usually contain complex background.

\textbf{DomainNet}~\cite{peng2019moment} is the largest benchmark so far which is formed by 0.6 million images from 345 classes. Six domains: Clipart (clp), Infograph (inf), Painting (pnt), Quickdraw (qdr), Real (rel) and Sketch (skt) are included, each of which contains a training set and a testing set. We train the domain adaptation network based on those training sets from each domain, and calculate accuracy on the testing sets.


\textbf{Baselines.} For single-target domain adaptation, we compare with several classical and state-of-the-art methods, namely, the classical DANN~\cite{ganin2015unsupervised}, MSTN~\cite{xie2018learning}, ADDA~\cite{tzeng2017adversarial}, MCD~\cite{saito2018maximum}, RTN~\cite{long2016unsupervised}, JAN~\cite{long2017deep}, CDAN+BSP~\cite{chen2019transferability}, CAN~\cite{kang2019contrastive}, MDD~\cite{zhang2019bridging}, DCAN~\cite{ge2020domain}, SHOT-IM~\cite{liang2020we}, MIMTFL~\cite{gao2020reducing}, and the most recent FixBi~\cite{na2021fixbi}, ATDOC~\cite{liang2021domain}, SCDA~\cite{li2021semantic}, AANet~\cite{xia2021adaptive}. For multi-target domain adaptation, we compare with three recent works MT-MTDA~\cite{nguyen2021unsupervised}, D-CGCT~\cite{roy2021curriculum}, and HGAN~\cite{yang2020heterogeneous}. Notice that the quantitative results are cited from published papers because we follow the same setting as those papers on the Office, Office-Home, and DomainNet benchmarks.

\textbf{Implementation Details} We employ the pre-trained ResNet-50 or ResNet-101 as the feature generator \textbf{G} like that in~\cite{liang2020we,chen2020adversariallearned,ge2020domain,xu2019larger,long2018conditional}. For the classifier \textbf{F}, we use a linear layer with the input and output dimensions being $(1024,k)$, where $k$ is the number of classes. The network is trained with the Stochastic Gradient Descent (SGD) optimizer with a momentum of 0.9. The learning rate is scheduled with the strategy in~\cite{ganin2016domain}, which is formulated as $\eta_{q}=\frac{\eta_{0}}{(1+aq)^{b}}$ where $q$ is the training progress linearly changed from 0 to 1, $a=10$, $b=0.75$, $\eta_{0}=0.01$ for \textbf{F} and the bottleneck layers, and $\eta_{0}=0.001$ for \textbf{G}. We use 48 as the training and testing batch size for all the experiments due to the limit of memory. The up-limit of training iterations for Office-31, Office-Home, and DomainNet are 3000, 5000, and 10000, respectively. The value of temperature $\tau$ in feature normalization is set to 20, following the results of~\cite{saito2019semi,ranjan2017l2}. Similar to the setting in~\cite{liang2020we}, we randomly run our method three times via PyTorch and report the average accuracy.

\subsection{Results}

    \begin{figure*}
    	\centering
    	\footnotesize
    	\setlength{\tabcolsep}{0.1mm}
    	\begin{tabular}{ccccccc}
    		\includegraphics[width=0.14\linewidth]{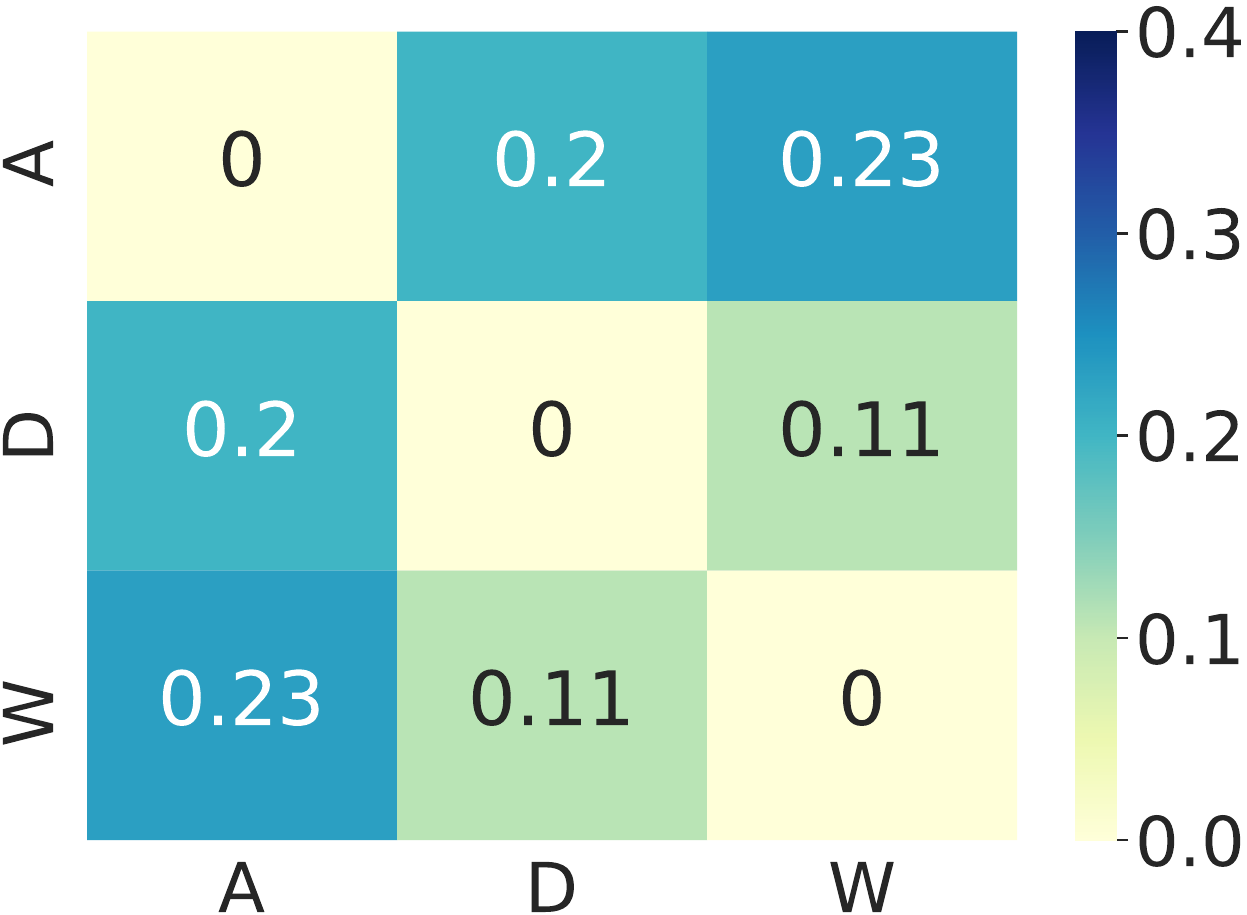}
    		& \includegraphics[width=0.14\linewidth]{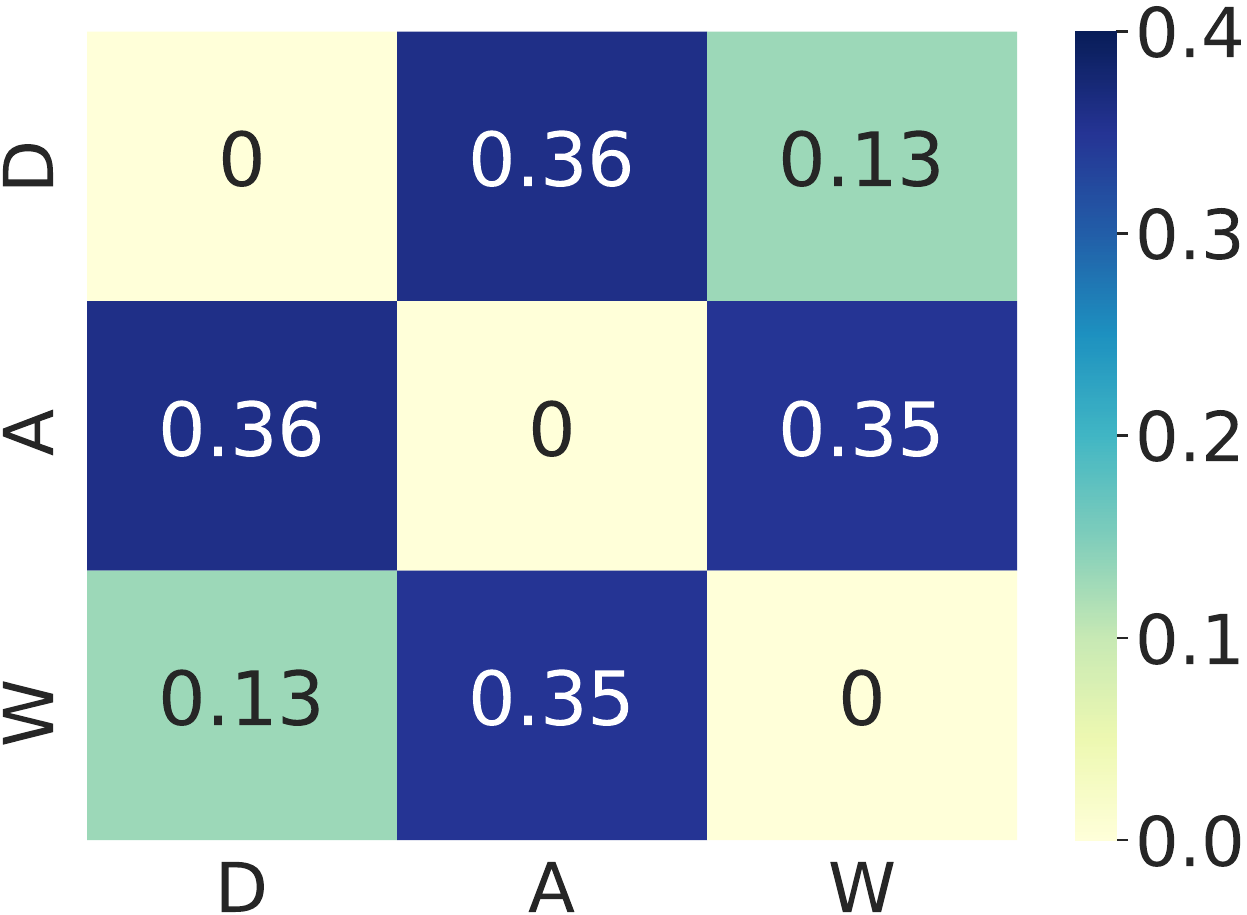}
    		& \includegraphics[width=0.14\linewidth]{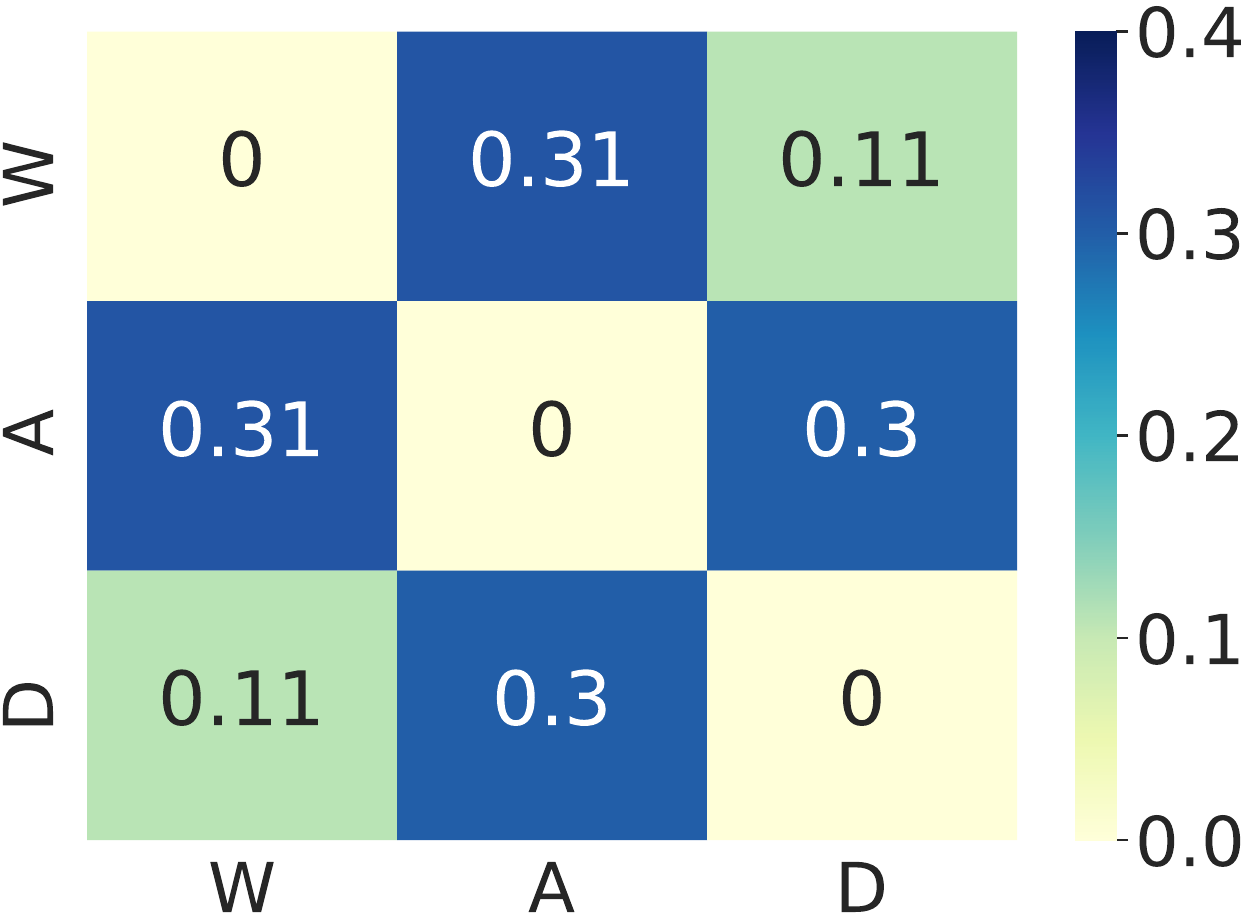}
    		& \includegraphics[width=0.14\linewidth]{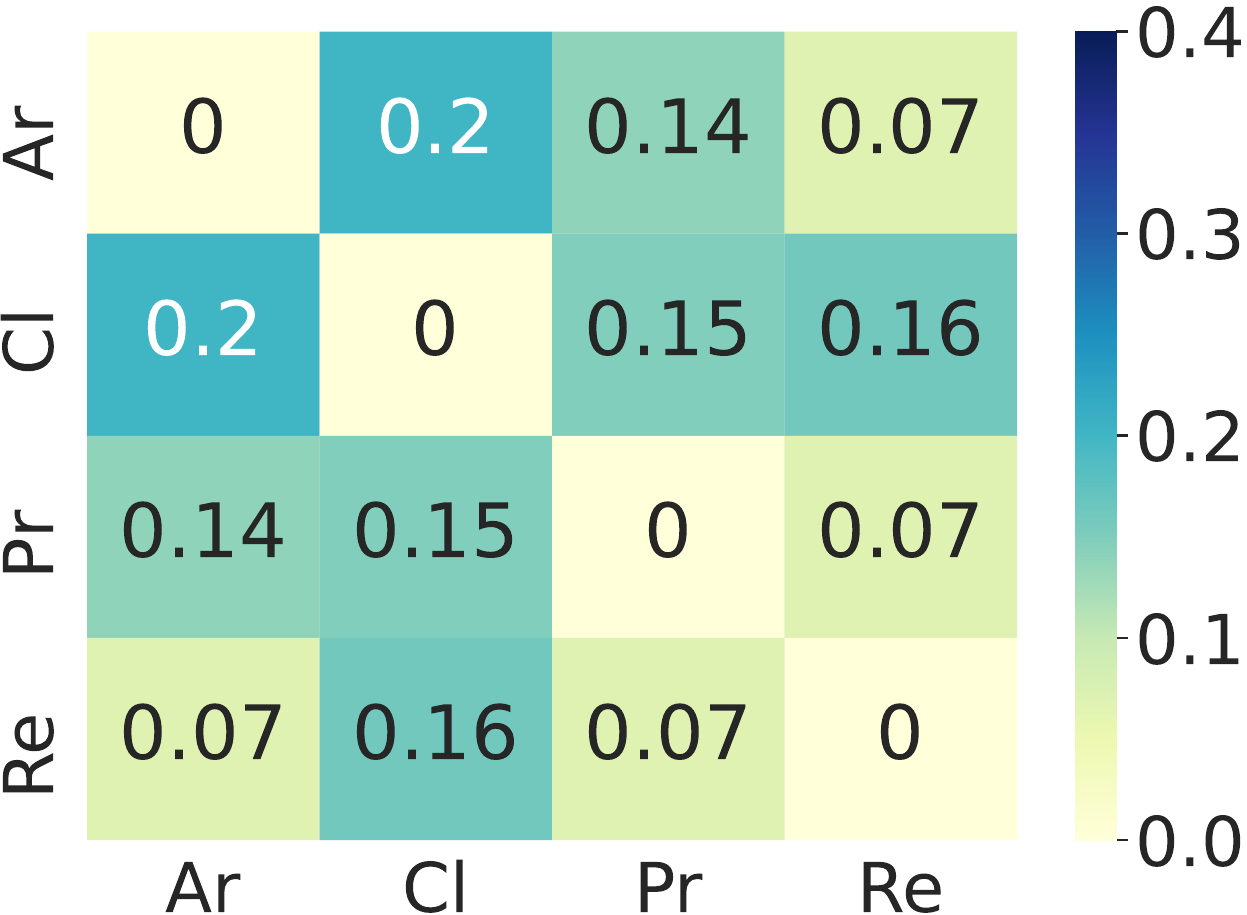}
    		& \includegraphics[width=0.14\linewidth]{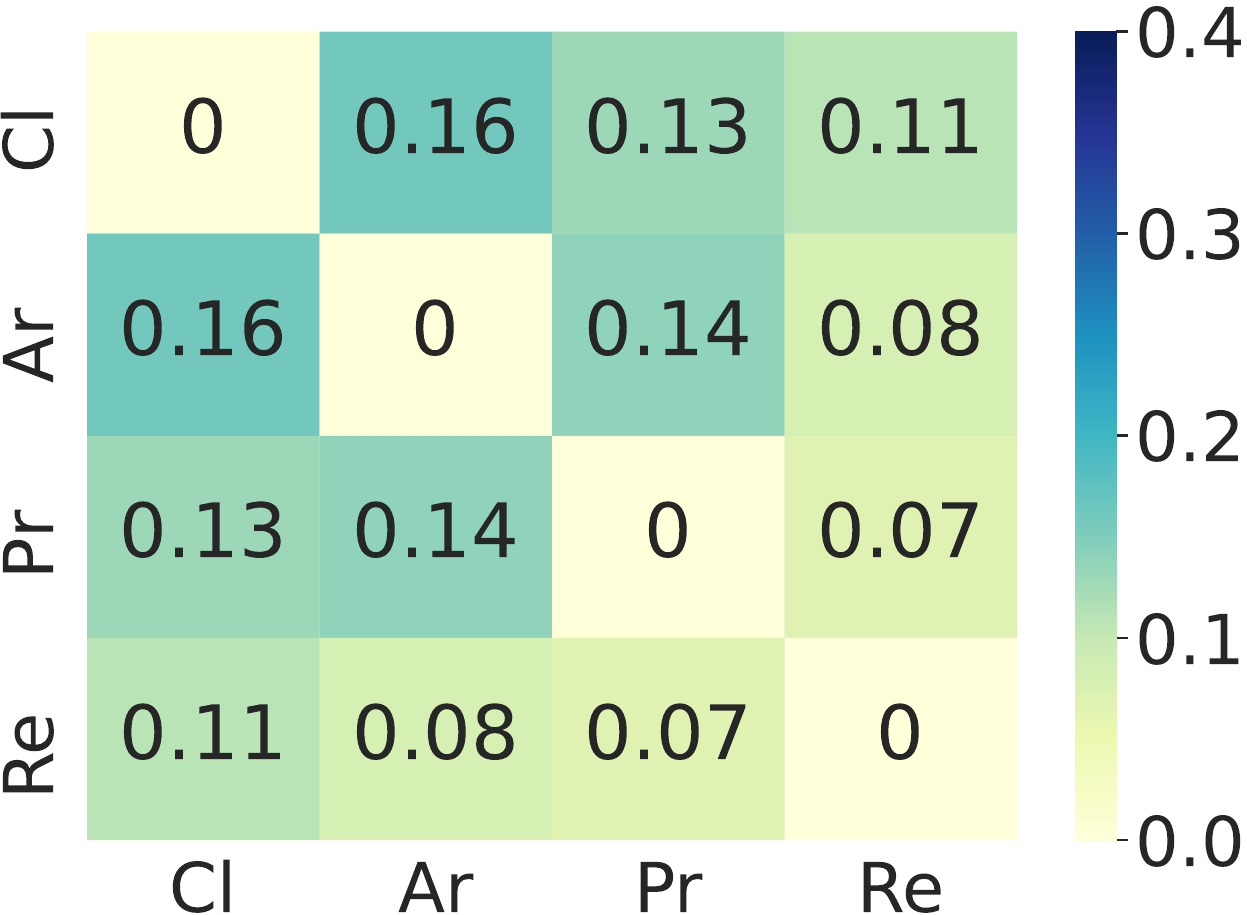}
    		& \includegraphics[width=0.14\linewidth]{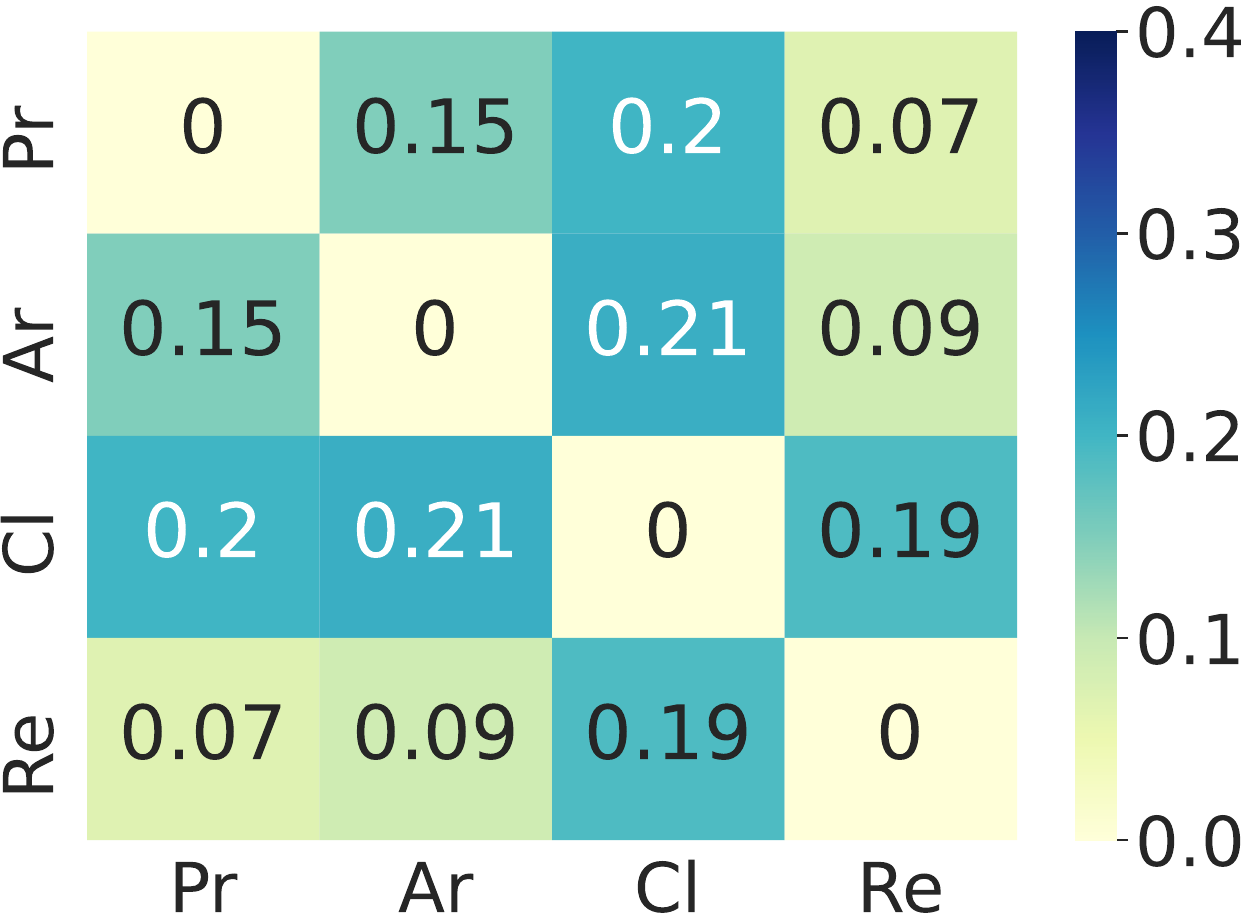}
    		& \includegraphics[width=0.14\linewidth]{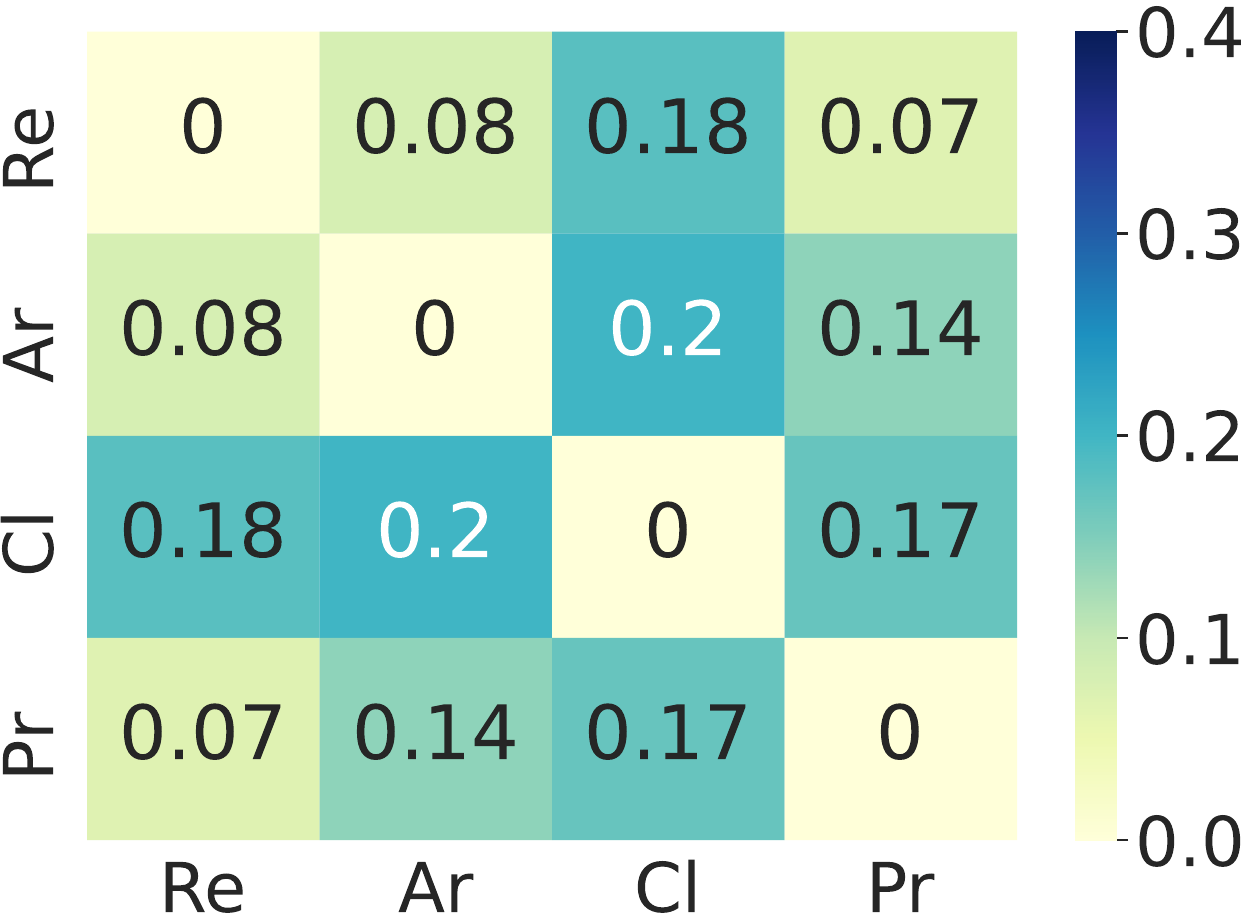} \\
    		(a) A &(b) D &(c) W &(d) Ar &(f) Cl &(g) Pr &(h) Re 
    	\end{tabular} 
    	\begin{tabular}{cccccc}
    		\includegraphics[width=0.16\linewidth]{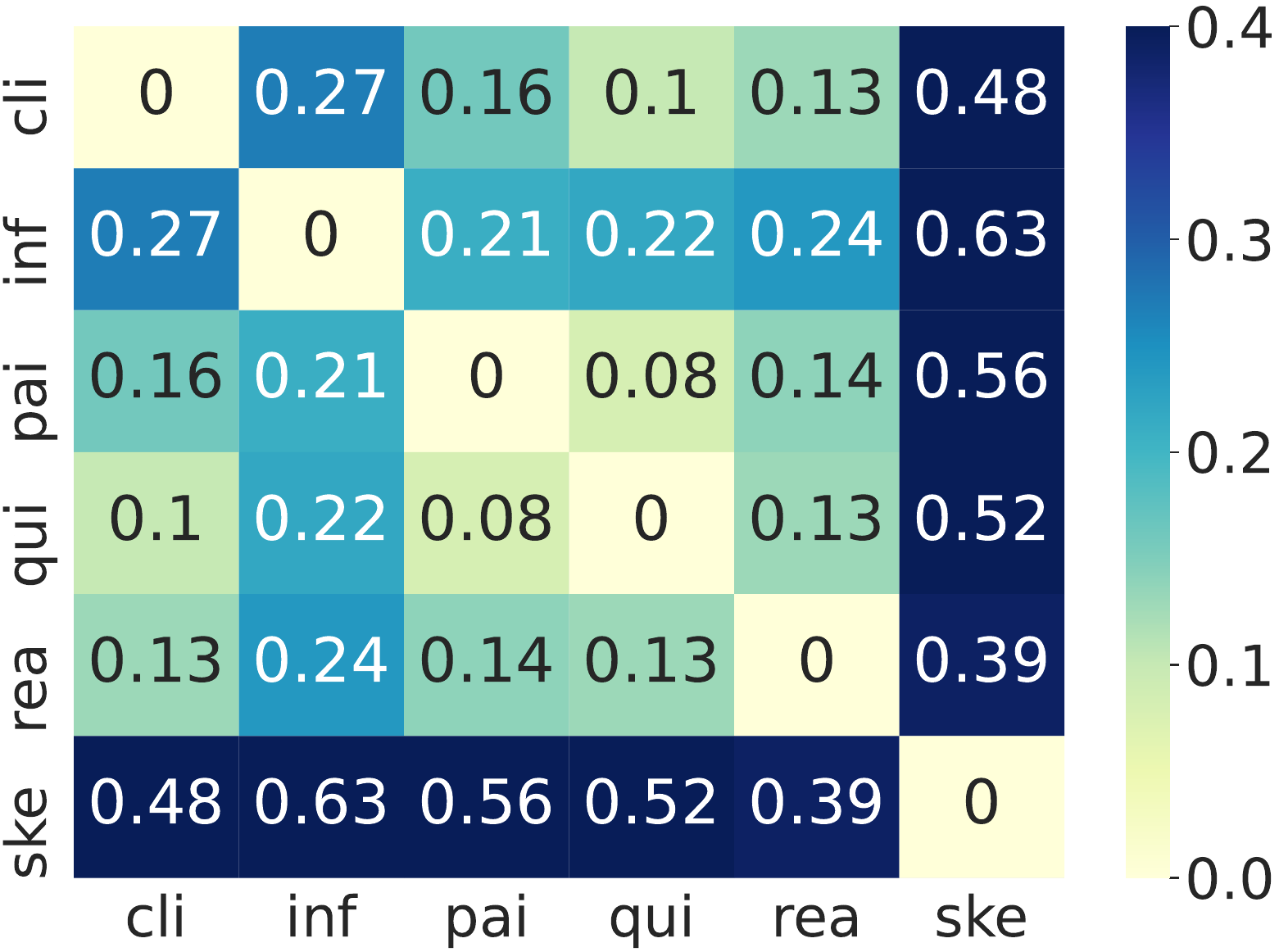}
    		& \includegraphics[width=0.16\linewidth]{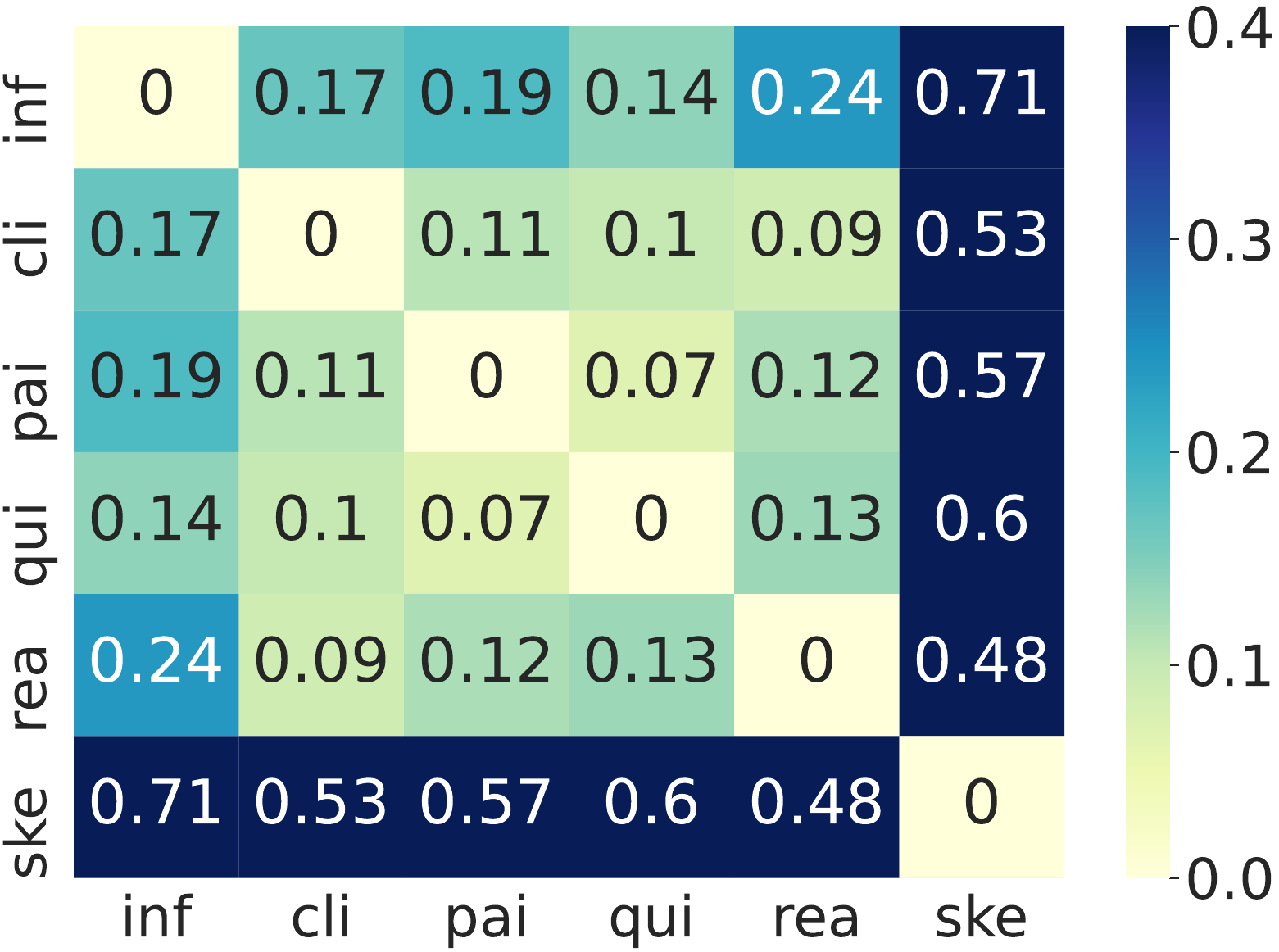}
    		& \includegraphics[width=0.16\linewidth]{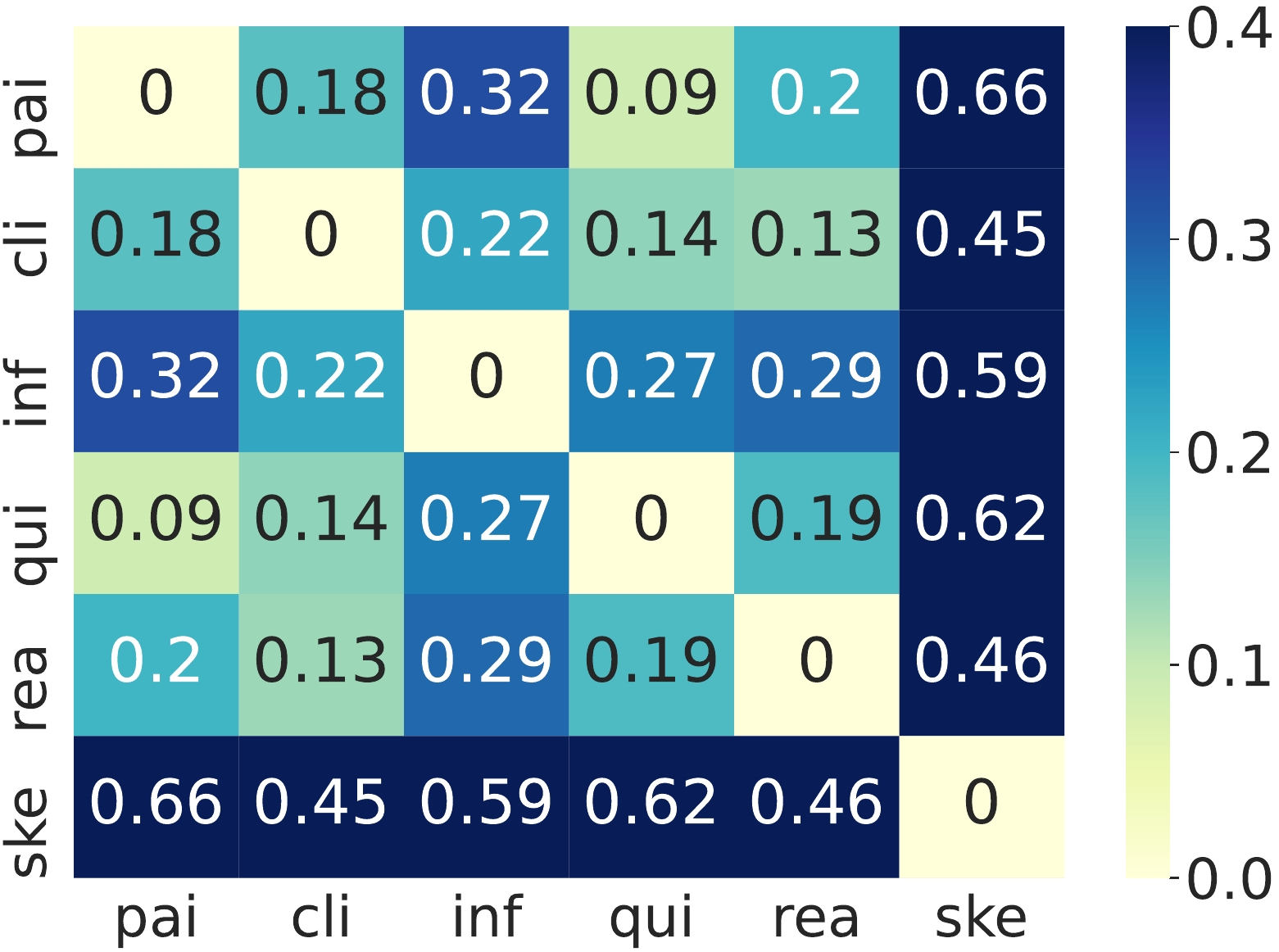}
    		& \includegraphics[width=0.16\linewidth]{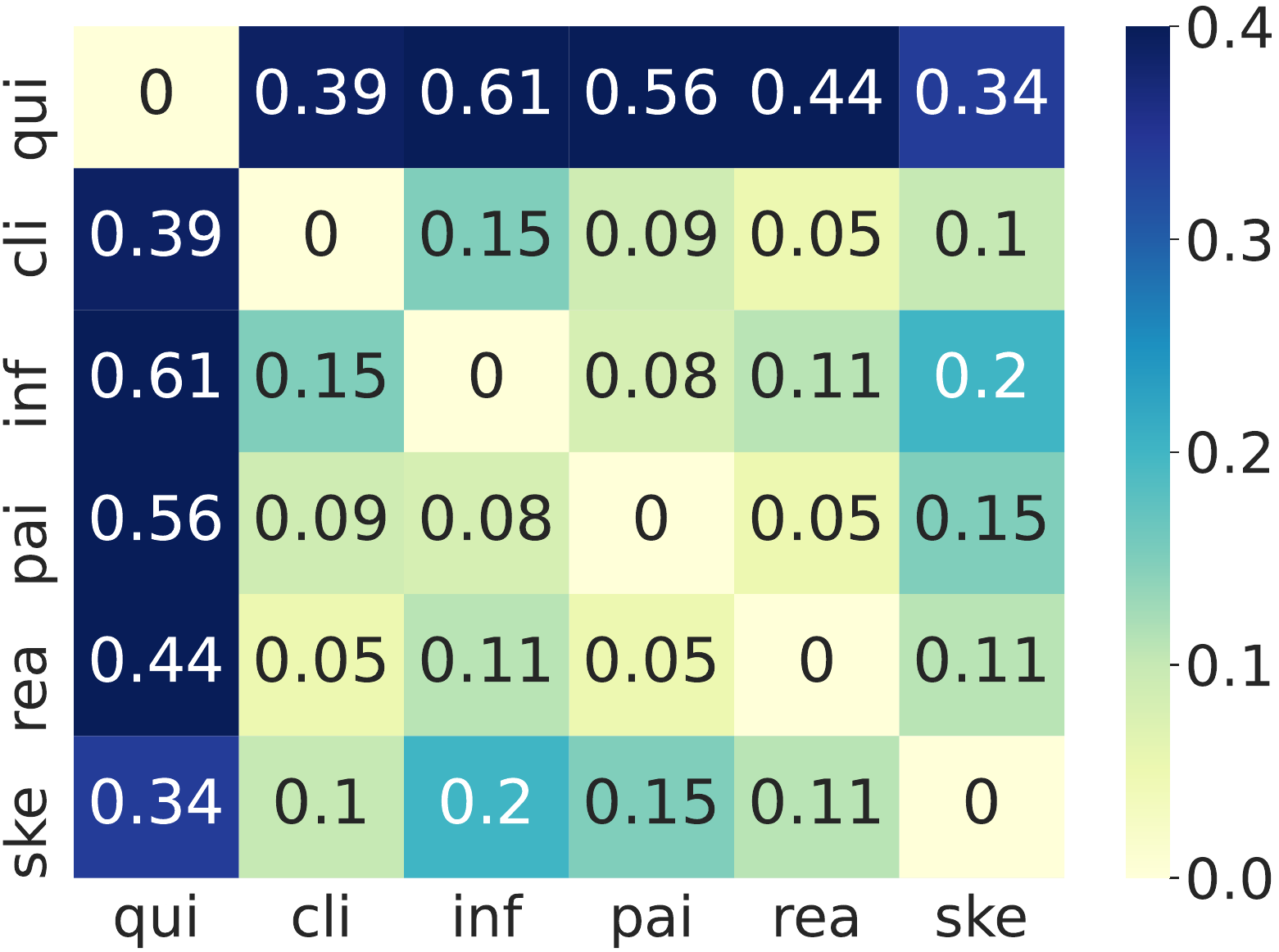}
    		& \includegraphics[width=0.16\linewidth]{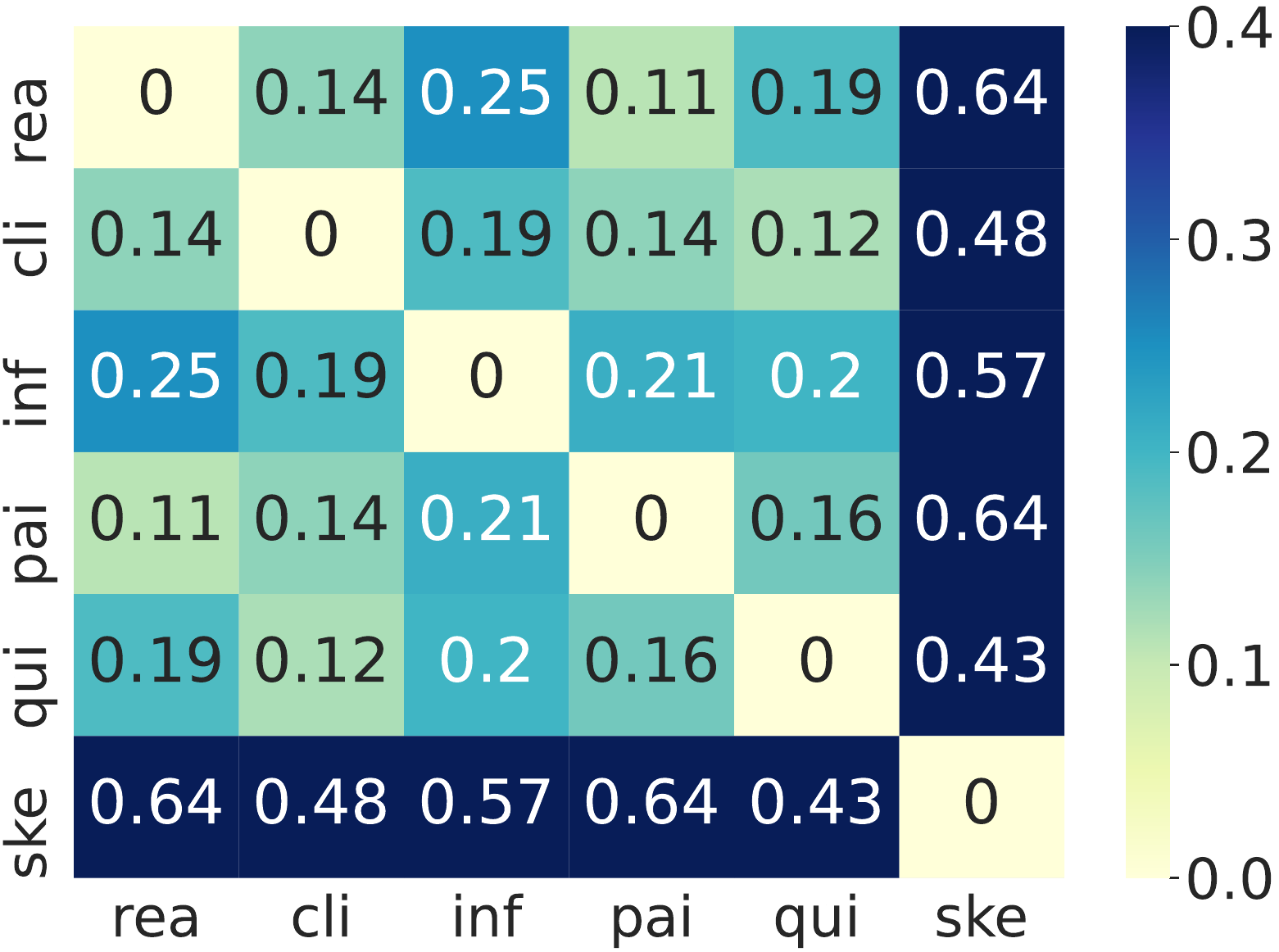}
    		& \includegraphics[width=0.16\linewidth]{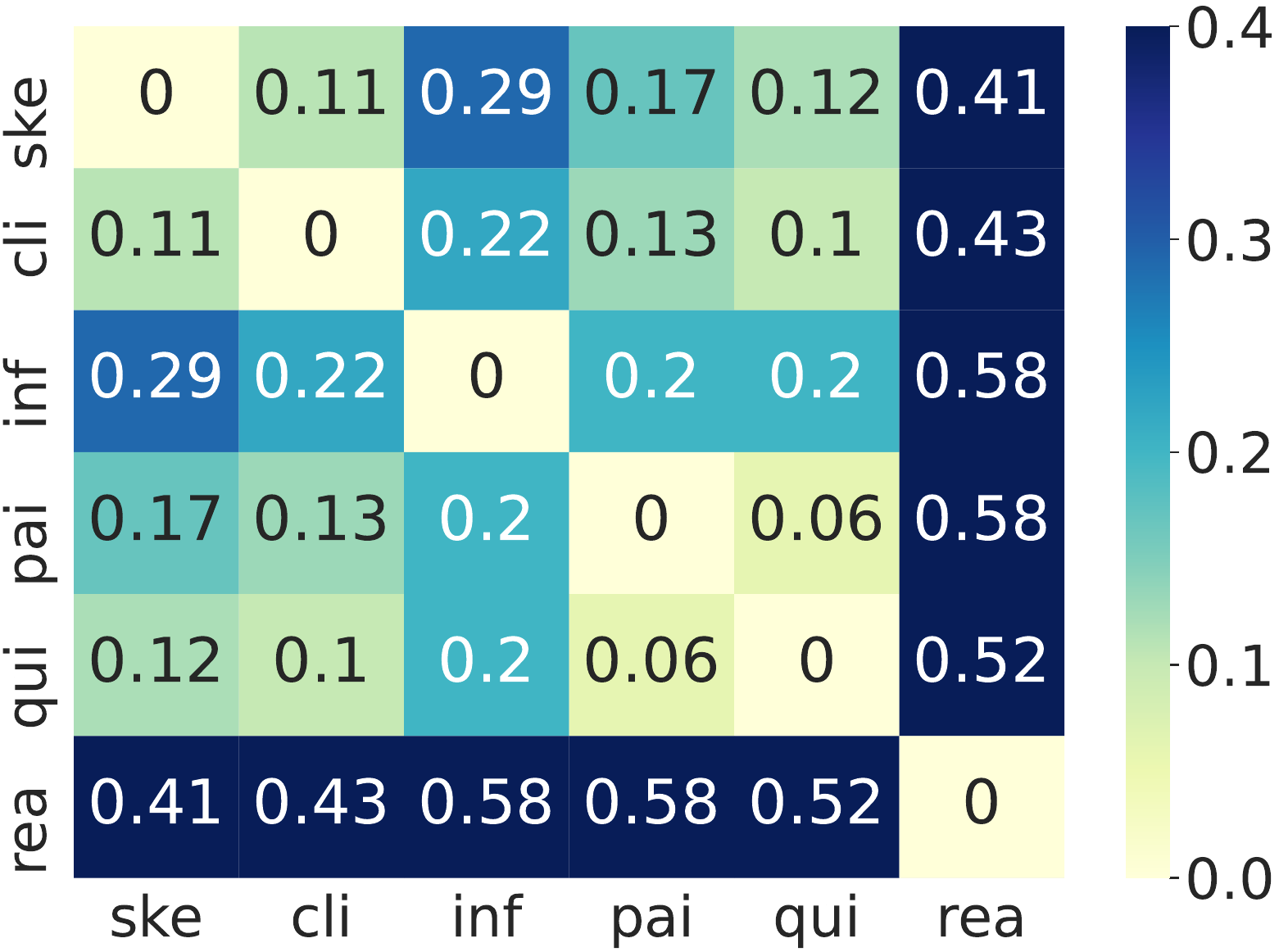}\\
    		(i) clp &(j) inf &(k) pnt &(l) qdr &(m) rel &(n) skt\\ 
    	\end{tabular}    	
    	\caption{The average of distance tensor $\mathcal{G}$ over different classes for Office-31, Office-Home, and DomainNet. Lighter color means smaller distance.}
    	\label{Fig:dis_matrix}
    \end{figure*} 

    \begin{figure}
    	\centering
    	\footnotesize
    	\begin{tabular}{c}
    		\includegraphics[width=0.95\linewidth]{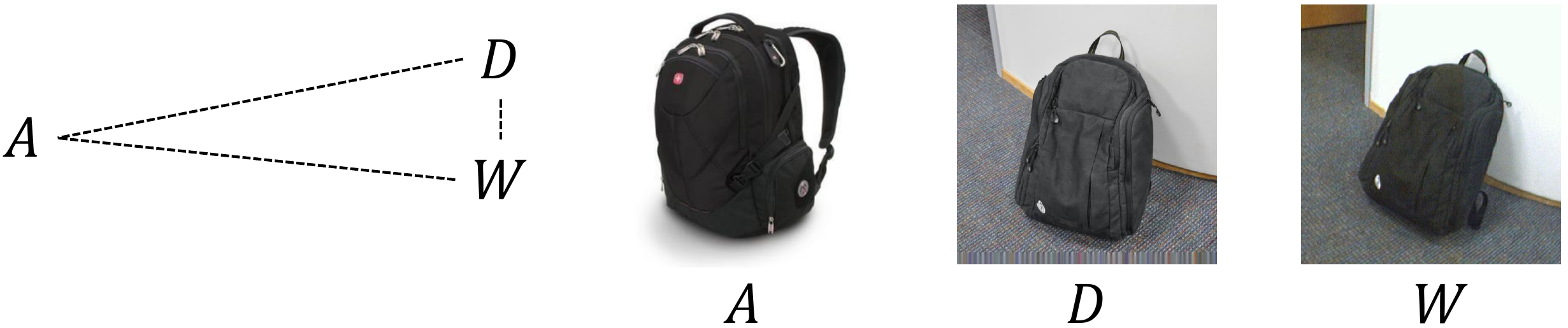} \\
    		(a) Office-31\\
    		\includegraphics[width=0.95\linewidth]{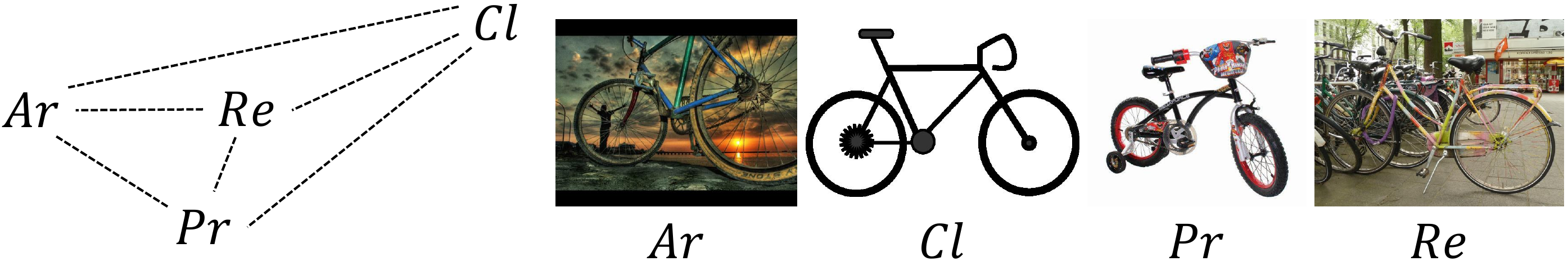} \\
    		(b) Office-Home
    	\end{tabular} 
    	\caption{The estimated topology relationship between domains and selected representative samples in Office-31 and Office-Home.}
    	\label{Fig:dis_demo}
    \end{figure} 

    \begin{table}[!ht]
        \centering
    	\caption{Classification accuracies (\%) on small-size Office-31 dataset with backbone ResNet-50.}
    	\label{tab:office31}
    	\scriptsize
    	\setlength{\tabcolsep}{0.3mm}
    	\renewcommand{\arraystretch}{1.1}
    	\begin{tabular}{c l| c c| c c| c c |c}
    	Category &Method (S$\xrightarrow{}$ T) & A$\xrightarrow{}$ D & A$\xrightarrow{}$ W & D$\xrightarrow{}$ A & D$\xrightarrow{}$ W & W$\xrightarrow{}$ A & W$\xrightarrow{}$ D & \textbf{Avg.} \\
    	\hline
    	\multirow{8}{*}{Single T.}    
    	&SHOT-IM~\cite{liang2020we}                  &90.6 &91.2 &72.5 &98.3 &71.4 &99.9 &87.3\\
    	&DCAN~\cite{ge2020domain}                    &90.7 &92.7 &73.5 &98.4 &73.4 &100. &88.1\\
    	&CDAN+BSP~\cite{chen2019transferability}     &93.0 &93.3 &73.6 &98.2 &72.6 &100. &88.5\\
    	&AANet~\cite{xia2021adaptive}                &94.5 &94.0 &76.7 &99.2 &76.1 &100. &90.1\\
    	&SCDA~\cite{li2021semantic}                  &94.6 &94.8 &77.5 &98.2 &76.4 &100. &90.3\\
    	&ATDOC~\cite{liang2021domain}                &95.4 &94.6 &77.5 &98.1 &77.0 &99.7 &90.4\\
     	&CAN~\cite{chen2019transferability}          &95.0 &94.5 &78.0 &99.1 &77.0 &99.8 &90.6\\
     	&FixBi~\cite{na2021fixbi}                    &95.0 &96.1 &78.7 &99.3 &79.4 &100. &91.4\\
    	\hline
    	\multirow{3}{*}{Multi T.}       
    	&MT-MTDA~\cite{nguyen2021unsupervised} &\multicolumn{2}{c|}{87.9} &\multicolumn{2}{c|}{83.7} &\multicolumn{2}{c|}{84.0} &85.2 \\
    	&HGAN~\cite{yang2020heterogeneous}           &87.8 &88.2 &71.4 &97.5 &69.9 &100.0 &85.8 \\
    	&D-CGCT~\cite{roy2021curriculum}       &\multicolumn{2}{c|}{93.4} &\multicolumn{2}{c|}{86.0} &\multicolumn{2}{c|}{87.1} & 88.8 \\ 
    	\hline
    	\multirow{2}{*}{Our} 
    	&SWISS(single)                 &94.8 &93.6 &76.5 &98.6 &77.1 &99.8 &90.1\\
    	&SWISS(multi)                  &94.7 &94.3 &77.1 &98.6 &77.0 &99.8 &90.3\\
    	\hline
        \end{tabular}
    \end{table}

    \begin{table*}[!ht]
        \centering
    	\caption{Classification accuracies (\%) on medium-size Office-Home dataset with backbone ResNet-50. }
    	\label{tab:office_home}
    	\scriptsize
    	\setlength{\tabcolsep}{0.9mm}
    	\renewcommand{\arraystretch}{1.1}
    	\begin{tabular}{c l | c c c | c c c | c c c | c c c |c}
    	\hline
    	Category &Method (S$\xrightarrow{}$ T) & Ar$\xrightarrow{}$ Cl & Ar$\xrightarrow{}$ Pr & Ar$\xrightarrow{}$ Re & Cl$\xrightarrow{}$ Ar & Cl$\xrightarrow{}$ Pr & Cl$\xrightarrow{}$ Re & Pr$\xrightarrow{}$ Ar & Pr$\xrightarrow{}$ Cl & Pr$\xrightarrow{}$ Re & Re$\xrightarrow{}$ Ar & Re$\xrightarrow{}$ Cl & Re$\xrightarrow{}$ Pr & \textbf{Avg.} \\
    	\hline
    	\multirow{8}{*}{Single T.}    
    	&CDAN+BSP~\cite{chen2019transferability}     &52.0 &68.6 &76.1 &58.0 &70.3 &70.2 &58.6 &50.2 &77.6 &72.2 &59.3 &81.9 &66.3\\
    	&SHOT-IM~\cite{liang2020we}                  &55.4 &76.6 &80.4 &66.9 &74.3 &75.4 &65.6 &54.8 &80.7 &73.7 &58.4 &83.4 &70.5\\
    	&DCAN~\cite{ge2020domain}                    &57.9 &76.2 &79.3 &67.3 &76.1 &75.6 &65.4 &56.0 &80.7 &74.2 &61.2 &84.2 &71.2\\
    	&FixBi~\cite{na2021fixbi}                    &58.1 &77.3 &80.4 &67.7 &79.5 &78.1 &65.8 &57.9 &81.7 &76.4 &62.9 &86.7 &72.7\\
    	&AANet~\cite{xia2021adaptive}                &58.4 &79.0 &82.4 &67.5 &79.3 &78.9 &68.0 &56.2 &82.9 &74.1 &60.5 &85.0 &72.8\\
    	&SCDA~\cite{li2021semantic}                  &60.7 &76.4 &82.8 &69.8 &77.5 &78.4 &68.9 &59.0 &82.7 &74.9 &61.8 &84.5 &73.1\\
    	&ATDOC~\cite{liang2021domain}                &60.2 &77.8 &82.2 &68.5 &78.6 &77.9 &68.4 &58.4 &83.1 &74.8 &61.5 &87.2 &73.2\\
    	\hline
    	\multirow{2}{*}{Multi T.}
    	&MT-MTDA~\cite{nguyen2021unsupervised} & &64.6 & & &66.4 & & &59.2 & & &67.1 & &64.3 \\
    	&D-CGCT~\cite{roy2021curriculum} & &70.5 & & &71.6 & & &66.0 & & &71.2 & &69.8 \\
    	\hline
    	\multirow{2}{*}{Our}
    	&SWISS(single)                 &60.7 &78.2 &81.9 &69.3 &77.9 &79.4 &69.5 &57.4 &82.6 &76.4 &59.8 &85.0 &73.2\\
    	&SWISS(multi)                  &61.2 &79.2 &81.8 &70.6 &78.5 &79.5 &70.7 &58.2 &82.6 &76.3 &59.7 &85.1 &73.6\\
    	\hline
        \end{tabular}
    \end{table*}
    
\textbf{Results on Office-31}
The classification results for the small-size Office-31 dataset are presented in Table 1. Although our methods are not as accurate as the recent single-target works FixBi~\cite{na2021fixbi} and CAN~\cite{chen2019transferability}, which significantly outperform other methods, both of our methods demonstrate similar performance to AANet~\cite{xia2021adaptive}, SCDA~\cite{li2021semantic}, and ATDOC~\cite{liang2021domain} at the second level. However, when compared to the most recent multi-target methods MT-MTDA~\cite{nguyen2021unsupervised}, HGAN~\cite{yang2020heterogeneous}, and D-CGCT~\cite{roy2021curriculum}, our SWISS(multi) exhibits a clear improvement of 1.5\% over their best performance. Analyzing SWISS(multi) in comparison to SWISS(single), we observe improvements in sub-tasks A$\xrightarrow{}$ W and D$\xrightarrow{}$ A, while no improvement is observed for the sub-tasks D$\xrightarrow{}$ W and W$\xrightarrow{}$ D. The reason behind this phenomenon can be understood from Fig.\ref{Fig:dis_matrix} (a)-(c), where it can be seen that the distance between A and D is slightly smaller than the distance between A and W in Fig.\ref{Fig:dis_matrix} (a). This implies that D can aid in domain adaptation between A and W, and similarly, W can enhance performance when adapting from D to A, as depicted in Fig.\ref{Fig:dis_matrix} (b). However, the distance from A to D or W is consistently larger than the distance between D and W, resulting in no improvement when adopting the domain between D and W. Furthermore, Fig.\ref{Fig:dis_demo}(a) illustrates the estimated topology relationship between domains and representative samples in the Office-31 dataset, revealing that samples from D and W exhibit high similarity to each other but are distinct from those in A. Consequently, when D or W serves as the source domain, samples from A are unable to provide valuable information to mitigate domain divergence. It is important to note that Office-31 is a small dataset and the D and W domains contain only 498 and 795 samples respectively, which means that a 1.0\% improvement corresponds to merely 5 to 8 accurately estimated samples. In the remainder of this section, we will demonstrate that our methods achieve significantly better performance than state-of-the-art methods in medium-size and large-size benchmarks.


\textbf{Results on Office-Home}
In the more challenging medium-size Office-Home dataset, our SWISS(multi) achieves state-of-the-art performance, surpassing other methods by 0.4\% at the second level (ATDOC~\cite{liang2021domain} and SCDA~\cite{li2021semantic}) and by 0.8\% at the third level (AANet~\cite{xia2021adaptive} and FixBi~\cite{na2021fixbi}), as shown in Table 2. Additionally, our SWISS(single) achieves an average accuracy of 73.1\%, which is on par with ATDOC~\cite{liang2021domain} and SCDA~\cite{li2021semantic}. When comparing our methods to others, we observe that our approach outperforms competing methods, particularly in challenging tasks like Ar$\xrightarrow{}$ Cl and Pr$\xrightarrow{}$ Ar. Comparing SWISS(multi) to SWISS(single), we observe clear improvements in both the average accuracy and most sub-tasks, such as Ar$\xrightarrow{}$ Cl, Ar$\xrightarrow{}$ Pr, and Cl$\xrightarrow{}$ Pr. The explanation for this can be found in Fig.\ref{Fig:dis_demo}(b), which reveals that Ar, Pr, and Cl form a triangle with Re inside. This indicates that when any of the domains Ar, Pr, or Cl is the source domain, Re can serve as a useful peer domain to enhance the performance of sub-tasks. Conversely, when Re is the source domain, none of the domains Ar, Pr, or Cl can act as a useful peer domain to each other, resulting in no improvement of SWISS(multi) over SWISS(single) in the Re$\xrightarrow{}$ Ar/Pr/Cl sub-tasks. The relationships between domains are further validated in Fig.\ref{Fig:dis_matrix}. For example, considering Ar as the source domain, we can observe from Fig.\ref{Fig:dis_matrix}(d) that the distance between Ar and Re ($dis_{Ar,Re}=0.07$) is smaller than the distance between Ar and Cl ($dis_{Ar,Cl}=0.2$), and the distance between Cl and Re ($dis_{Cl,Re}=0.16$) is also smaller than $dis_{Ar,Cl}$. Thus, both criteria in Sec.\ref{sec:peerscaffolding} are satisfied, allowing domain Re to be utilized to aid the adaptation process of Ar$\xrightarrow{}$ Cl. Comparing our approach to the multi-target methods MT-MTDA~\cite{nguyen2021unsupervised} and D-CGCT~\cite{roy2021curriculum}, we observe a significant improvement of 3.8\% with our SWISS(multi).

\textbf{Results on DomainNet}
In the most challenging large-size DomainNet dataset, both our SWISS(single) and SWISS(multi) outperform state-of-the-art methods by a significant margin of approximately 2.0\%. The details of sub-task accuracies are presented in Table 3, revealing that for the very challenging sub-tasks with accuracies below 20.0\%, our SWISS(single) and SWISS(multi) exhibit a slight decrement of 1.0\% and 0.6\% on average compared to SCDA~\cite{li2021semantic}. However, for sub-tasks with accuracies above 20.0\%, our SWISS(single) and SWISS(multi) showcase a notable improvement of 3.1\% and 3.3\% on average, respectively, compared to SCDA~\cite{li2021semantic}. One possible reason for this phenomenon is that the predicted labels of target samples are heavily influenced by errors in those very challenging sub-tasks. As a result, both strong supervision and weak supervision struggle to provide useful information for enhancing performance. However, for moderately challenging tasks, both of our methods exhibit significant improvements over SCDA~\cite{li2021semantic} with a substantial margin.
\tab~\ref{tab:domainnet_average} provides a comparison with additional methods for each source$\xrightarrow{}$rest direction. From the table, several observations can be made: 1) The state-of-the-art multi-target method D-CGCT~\cite{roy2021curriculum} achieves a 1.1\% improvement over the state-of-the-art single-target method SCDA~\cite{li2021semantic}. 2) Our SWISS(single) and SWISS(multi) outperform D-CGCT~\cite{roy2021curriculum} with a moderate margin of around 1.0\%. Furthermore, our SWISS(multi) achieves the best scores in four out of six tasks among all the methods, demonstrating the effectiveness of our approach.

    \begin{table*}[!ht]
        \centering
    	\caption{Classification accuracies (\%) on large-size DomainNet dataset with backbone ResNet-101. The classification accuracy are reported for each source$\xrightarrow{}$target sub-tasks with source domain being indicated in the rows and target domain being indicated by columns.}
    	\label{tab:domainnet_detail}
    	\scriptsize
    	\setlength{\tabcolsep}{0.6mm}
    	\renewcommand{\arraystretch}{1.1}
    	\begin{tabular}{|c | c c c c c c c| c | c c c c c c c| c | c c c c c c c|}
    	\hline
    	ADDA\cite{tzeng2017adversarial} &clp &inf &pnt &qdr &rel &skt &Avg. &DANN~\cite{ganin2015unsupervised} &clp &inf &pnt &qdr &rel &skt &Avg. &MIMTFL~\cite{gao2020reducing} &clp &inf &pnt &qdr &rel &skt &Avg. \\
    	\hline \hline
    	clp &- &11.2 &24.1 &3.2 &41.9 &30.7 &22.2 &clp &- &15.5 &34.8 &9.5 &50.8 &41.4 &30.4 &clp &- &15.1 &35.6 &10.7 &51.5 &43.1 &31.2 \\
    	inf &19.1 &- &16.4 &3.2 &26.9 &14.6 &16.0 &inf &31.8 &- &30.2 &3.8 &44.8 &25.7 &27.3 &inf &32.1 &- &31.0 &2.9 &48.5 &31.0 &29.1 \\
    	pnt &31.2 &9.5 &- &8.4 &39.1 &25.4 &22.7 &pnt &39.6 &15.1 &- &5.5 &54.6 &35.1 &30.0 &pnt &40.1 &14.7 &- &4.2 &55.4 &36.8 &30.2 \\
    	qdr &15.7 &2.6 &5.4 &- &9.9 &11.9 &9.1   &qdr &11.8 &2.0 &4.4 &- &9.8 &8.4 &7.3     &qdr &18.8 &3.1 &5.0 &- &16.0 &13.8 &11.3 \\
    	rel &39.5 &14.5 &29.1 &12.1 &- &25.7 &24.2 &rel &47.5 &17.9 &47.0 &6.3 &- &37.3 &31.2 &rel &48.5 &19.0 &47.6 &5.8 &- &39.4 &32.1 \\
    	skt &35.3 &8.9 &25.2 &14.9 &37.6 &- &25.4 &skt &47.9 &13.9 &34.5 &10.4 &46.8 &- &30.7 &skt &51.7 &16.5 &40.3 &12.3 &53.5 &- &34.9 \\
    	Avg. &28.2 &9.3 &20.1 &8.4 &31.1 &21.7 &19.8 &Avg. &35.7 &12.9 &30.2 &7.1 &41.4 &29.6 &26.1 &Avg. &38.2 &13.7 &31.9 &7.2 &45.0 &32.8 &28.1 \\
    	
    	\hline \hline
    	ResNet-101~\cite{he2016deep} &clp &inf &pnt &qdr &rel &skt &Avg. &CDAN~\cite{long2018conditional} &clp &inf &pnt &qdr &rel &skt &Avg. &MDD~\cite{zhang2019bridging} &clp &inf &pnt &qdr &rel &skt &Avg. \\
    	\hline \hline
    	clp &- &19.3 &37.5 &11.1 &52.2 &41.0 &32.2 &clp &- &20.4 &36.6 &9.0 &50.7 &42.3 &31.8 &clp &- &20.5 &40.7 &6.2 &52.5 &42.1 &32.4 \\
    	inf &30.2 &- &31.2 &3.6 &44.0 &27.9 &27.4 &inf &27.5 &- &25.7 &1.8 &34.7 &20.1 &22.0 &inf &33.0 &- &33.8 &2.6 &46.2 &24.5 &28.0 \\
    	pnt &39.6 &18.7 &- &4.9 &54.5 &36.3 &30.8 &pnt &42.6 &20.0 &- &2.5 &55.6 &38.5 &31.8 &pnt &43.7 &20.4 &- &2.8 &51.2 &41.7 &32.0 \\
    	qdr &7.0 &0.9 &1.4 &- &4.1 &8.3 &4.3 &qdr &21.0 &4.5 &8.1 &- &14.3 &15.7 &12.7 &qdr &18.4 &3.0 &8.1 &- &12.9 &11.8 &10.8 \\
    	rel &48.4 &22.2 &49.4 &6.4 &- &38.8 &33.0 &rel &51.9 &23.3 &50.4 &5.4 &- &41.4 &34.5 &rel &52.8 &21.6 &47.8 &4.2 &- &41.2 &33.5\\
    	skt &46.9 &15.4 &37.0 &10.9 &47.0 &- &31.4 &skt &50.8 &20.3 &43.0 &2.9 &50.8 &- &33.6 &skt &54.3 &17.5 &43.1 &5.7 &54.2 &- &35.0\\
    	Avg. &34.4 &15.3 &31.3 &7.4 &40.4 &30.5 &26.6 &Avg. &38.8 &17.7 &32.8 &4.3 &41.2 &31.6 &27.7 &Avg. &40.4 &16.6 &34.7 &4.3 &43.4 &32.3 &28.6 \\
    	
    	\hline \hline
    	SCDA~\cite{li2021semantic} &clp &inf &pnt &qdr &rel &skt &Avg. &SWISS(single) &clp &inf &pnt &qdr &rel &skt &Avg. &SWISS(multi) &clp &inf &pnt &qdr &rel &skt &Avg. \\
    	\hline \hline
    	clp &- &20.4 &43.3 &15.2 &59.3 &46.5 &36.9 &clp &-    &19.6 &46.8 &15.9 &62.8 &49.3 &38.9    &clp &-    &19.5 &47.2 &15.8 &63.0 &49.5 &39.0 \\
    	inf &32.7 &- &34.5 &6.3 &47.6 &29.2 &30.1  &inf &39.5 &-    &39.7 &3.9 &53.4 &34.6 &34.2     &inf &40.4 &-    &40.2 &4.0  &53.6 &34.2 &34.5 \\
    	pnt &46.4 &19.9 &- &8.1 &58.8 &42.9 &35.2  &pnt &50.0 &19.2 &-    &5.4 &60.1 &45.0 &36.0     &pnt &50.0 &19.6 &-    &6.9  &60.1 &45.4 &36.4 \\
    	qdr &31.1 &6.6 &18.0 &- &28.8 &22.0 &21.3  &qdr &34.9 &5.9  &18.2 &-   &30.4 &24.7 &22.8     &qdr &35.5 &6.8  &19.5 &-    &30.5 &24.7 &23.4 \\
    	rel &55.5 &23.7 &52.9 &9.5 &- &45.2 &37.4  &rel &58.3 &21.8 &54.3 &6.9 &-    &46.9 &37.6     &rel &58.3 &22.0 &54.6 &6.9  &-    &47.3 &37.8 \\
    	skt &55.8 &20.1 &46.5 &15.0 &56.7 &- &38.8 &skt &60.7 &20.8 &49.4 &15.0 &61.2 &-   &41.4     &skt &60.8 &20.7 &49.6 &14.8 &61.5 &-    &41.5 \\
    	Avg. &44.3 &18.1 &39.0 &10.8 &50.2 &37.2 &33.3 &Avg. &48.7 &17.5 &41.7 &9.4 &53.6 &40.1 &35.2 &Avg. &49.0 &17.7 &42.2 &9.7 &53.7 &40.2 &35.4 \\
    	\hline
        \end{tabular}
    \end{table*}

    \begin{table}[!ht]
        \centering
    	\caption{Classification accuracies (\%) on large-size DomainNet dataset with backbone ResNet-101. The classification accuracy are reported for each source$\xrightarrow{}$rest direction, with each source domain being indicated in the columns.}
    	\label{tab:domainnet_average}
    	\scriptsize
    	\setlength{\tabcolsep}{2.0mm}
    	\renewcommand{\arraystretch}{1.1}
    	\begin{tabular}{c c c c c c c c }
    	\hline
    	\multirow{2}{*}{Model} 
    	&\multicolumn{7}{c}{DomainNet} \\
    	\cline{2-8}
    	&clp &inf &pnt &qdr &rel &skt &Avg. \\
    	\hline
        Source train &25.6 &16.8 &25.8 &9.2 &20.6 &22.3 &20.1\\
        SE~\cite{french2017self} &21.3 &8.5 &14.5 &13.8 &16.0 &19.7 &15.6 \\
        MCD~\cite{saito2018maximum} &25.1 &19.1 &27.0 &10.4 &20.2 &22.5 &20.7 \\
        DADA~\cite{peng2019domain} &26.1 &20.0 &26.5 &12.9 &20.7 &22.8 &21.5 \\
        CDAN~\cite{long2018conditional} &31.6 &27.1 &31.8 &12.5 &33.2 &35.8 &28.7 \\
        MCC~\cite{jin2020minimum} &33.6 &30.0 &32.4 &13.5 &28.0 &35.3 &28.8\\
        SCDA~\cite{li2021semantic} &36.9 &30.1 &35.2 &21.3 &37.4 &38.8 &33.3\\
        D-CGCT~\cite{roy2021curriculum} &37.0 &32.2 &37.3 &19.3 &39.8 &40.8 &34.4\\
        \hline
        SWISS(multi)  &\textbf{39.0} &\textbf{34.5} &36.4 &\textbf{23.4} &37.8 &\textbf{41.5} &\textbf{35.4}\\
        \hline
        \end{tabular}
    \end{table}

\subsection{Model Analysis and Discussions}
\begin{figure*}
    	\centering
    	\footnotesize
    	\setlength{\tabcolsep}{0.1mm}
    	\begin{tabular}{cccc}
    		\includegraphics[width=0.24\linewidth]{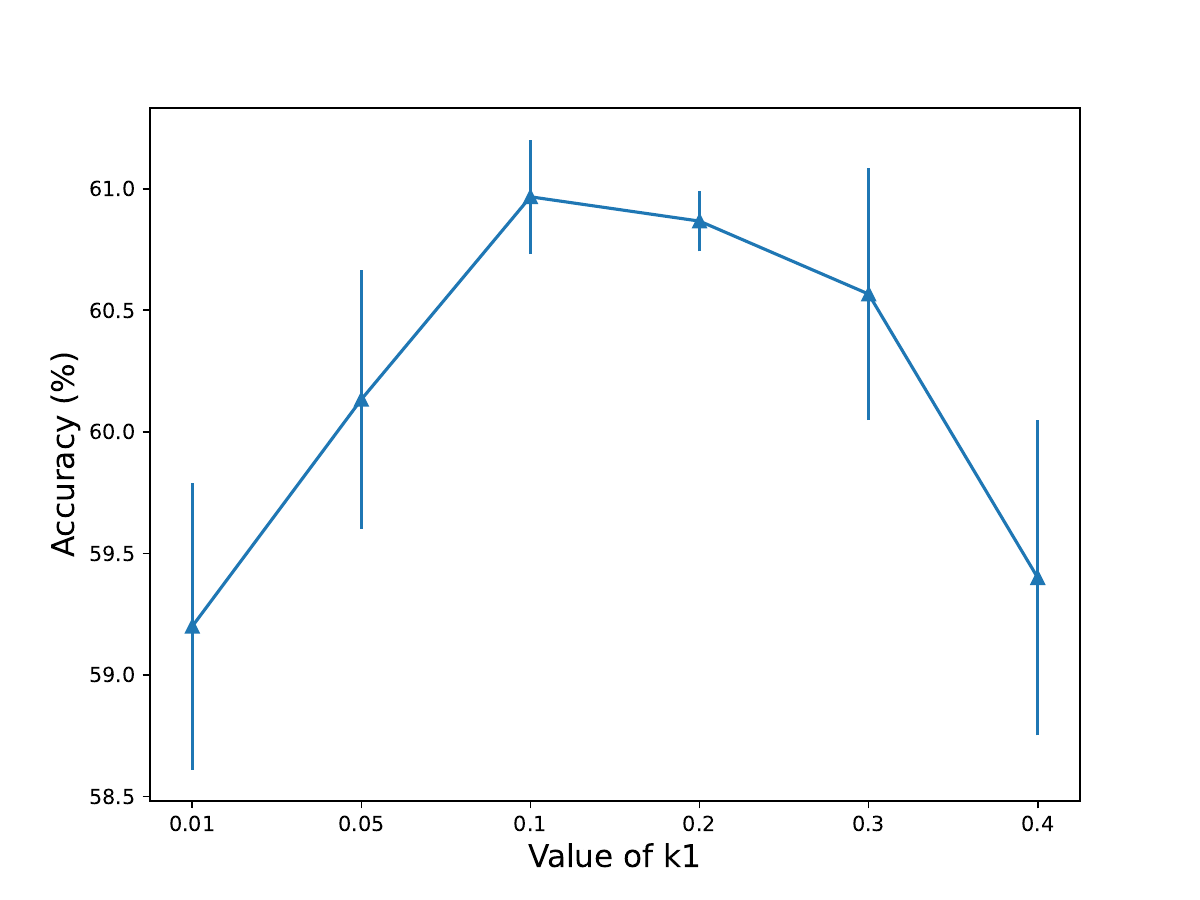}
    		& \includegraphics[width=0.24\linewidth]{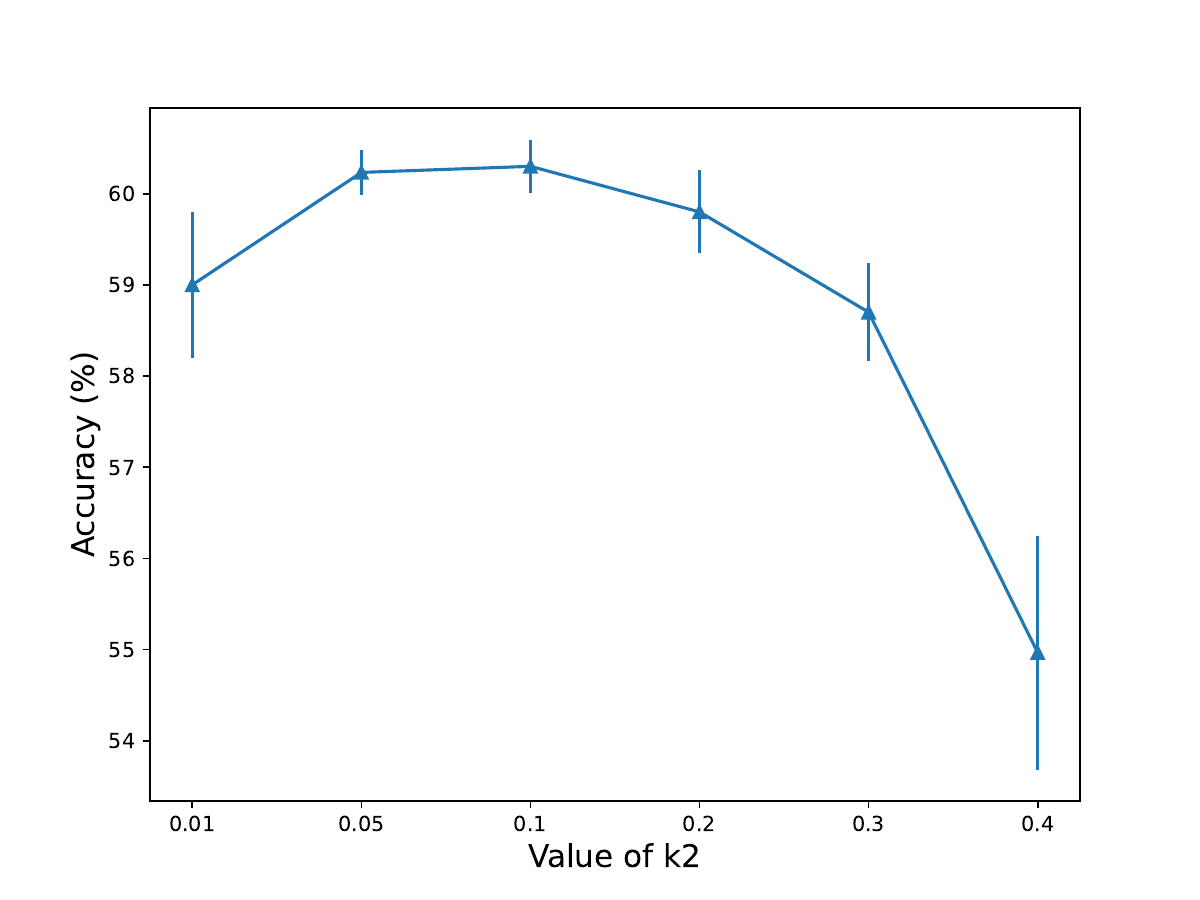}
    		& \includegraphics[width=0.24\linewidth]{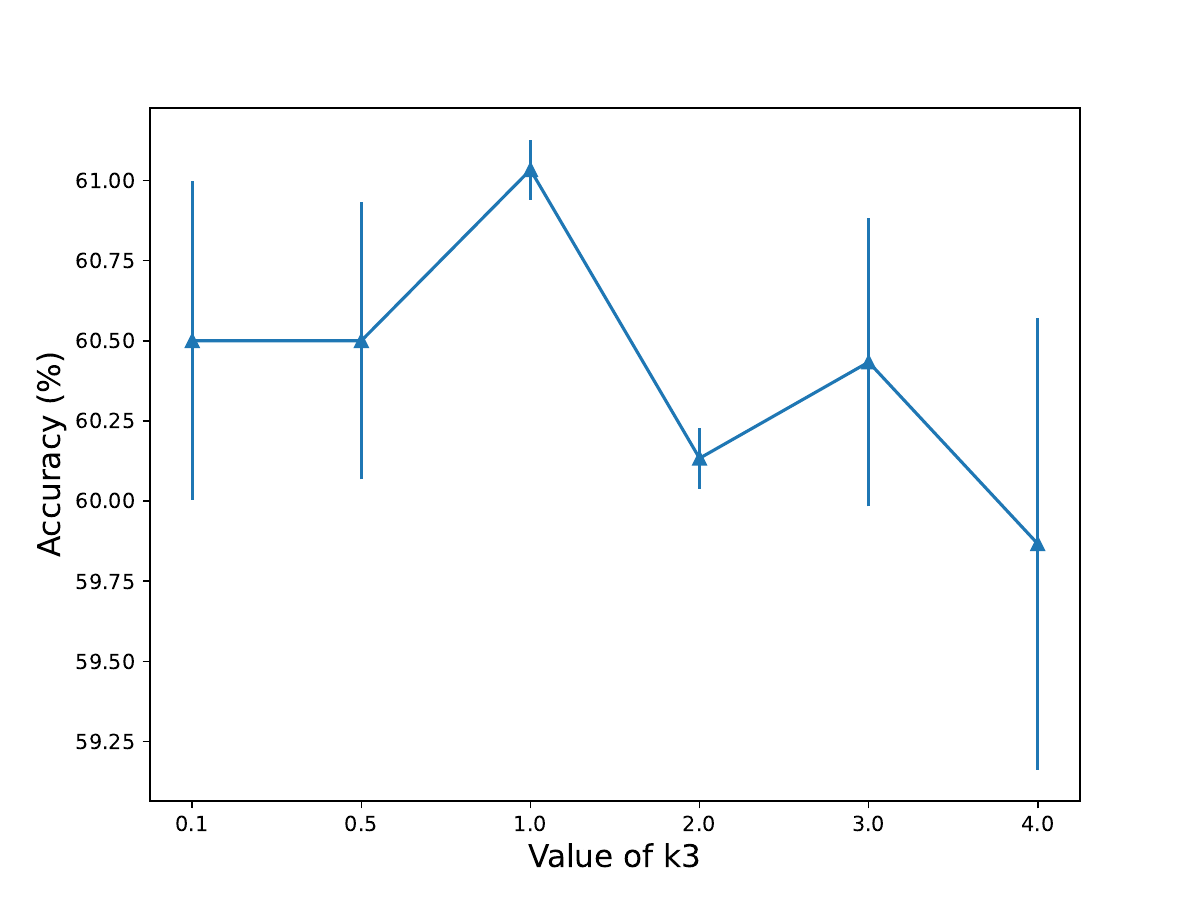}
    		& \includegraphics[width=0.24\linewidth]{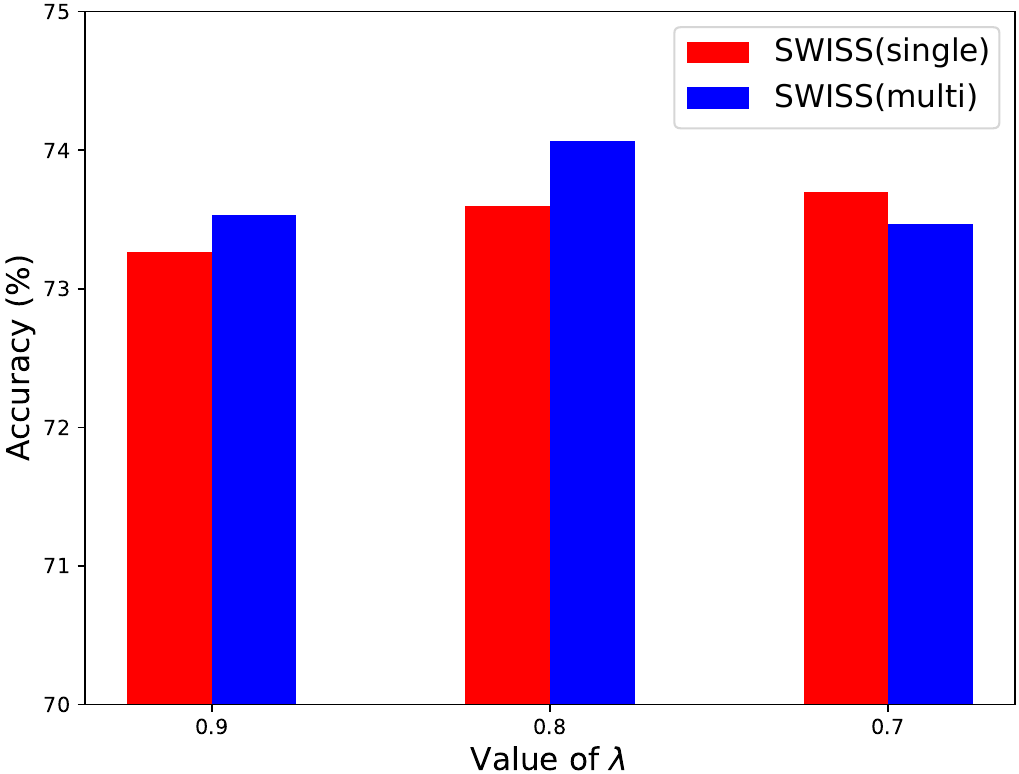} \\
    		(a)  &(b) &(c) &(d) \\
    	\end{tabular} 
    	\caption{Parameter sensitivity analysis with (a) k1=\{0.01, 0.05, 0.1, 0.2, 0.3, 0.4\}, (b) k2=\{0.01, 0.05, 0.1, 0.2, 0.3, 0.4\}, (c) k3=\{0.1, 0.5, 1.0, 2.0, 3.0, 4.0\}, and (d) $\lambda$=\{0.9, 0.8, 0.7\}. For (a)-(c), we evaluate in task Ar$\xrightarrow{}$ Cl. For (d), we show the average accuracies of Ar$\xrightarrow{}$ \{Cl, Pr, Re\} in both single-target and multi-target settings.}
    	\label{Fig:para_sensitivity}
    \end{figure*}

\textbf{Parameter sensitivity.}
To analyze the impact of the hyper-parameters in our method, we evaluate the accuracies by varying each parameter while keeping others at their default settings. The four hyper-parameters are the weights k1, k2, k3 of the loss terms $L_{IM}$, $L_{ALL}$, $L_{SW}$ and the threshold $\lambda$ for the prediction probability in \equ~\eqref{equ:weak_update} and \equ~\eqref{equ:l_all}. We conduct experiments with different values for each parameter: k1=\{0.01, 0.05, 0.1, 0.2, 0.3, 0.4\}, k2=\{0.01, 0.05, 0.1, 0.2, 0.3, 0.4\}, k3=\{0.1, 0.5, 1.0, 2.0, 3.0, 4.0\}, and $\lambda$=\{0.9, 0.8, 0.7\}. From Fig.~\ref{Fig:para_sensitivity}(a)-(c), we observe that the combination of k1=0.1, k2=0.05, and k3=1.0 leads to the highest accuracy and exhibits relatively smaller sensitivity to parameter changes. Regarding the threshold $\lambda$, we find that setting $\lambda$=0.8 yields the best performance. Specifically, in both single-target and multi-target settings, $\lambda$=0.9 performs worse than $\lambda$=0.8, while $\lambda$=0.7 performs slightly better than $\lambda$=0.8 in the single-target setting but significantly worse in the multi-target setting. Therefore, we choose the values k1=0.1, k2=0.05, k3=1.0, and $\lambda$=0.8 for all our experiments.

    \begin{table}[!ht]
        \centering
    	\caption{Classification accuracies (\%) on Office-Home with backbone ResNet-50 under different configurations of loss functions. The average accuracies are reported for each source in the columns which are evaluated on the single-target setting.}
    	\label{tab:ablation_loss}
    	\scriptsize
    	\begin{tabular}{l c c c c c}
    	\hline
    	Method & Ar &Cl &Pr &Re &\textbf{Avg.}\\
    	\hline
    	ResNet50~\cite{he2016deep} &47.6 &41.8 &43.4 &51.7 &46.1\\
    	\hline
    	$+L_{IM}$                      &71.5 &73.4 &66.9 &71.5 &70.8\\
    	$+L_{IM}+L_{ALL}$              &73.2 &73.7 &68.5 &72.5 &72.0\\
    	$+L_{IM}+L_{SW}$               &72.3 &74.4 &67.7 &73.2 &71.9\\
    	$+L_{ALL}+L_{SW}$              &72.6 &73.7 &68.6 &73.5 &72.1\\
    	$+L_{IM}+L_{ALL}+L_{SW}$       &73.6 &75.5 &69.8 &73.7 &73.2\\
    	\hline
        \end{tabular}
    \end{table}

    \begin{table}[!ht]
        \centering
    	\caption{Classification accuracies (\%) on Office-Home with backbone ResNet-50 under different configurations of network component.}
    	\label{tab:ablation_component}
    	\scriptsize
    	\setlength{\tabcolsep}{1.0mm}
    	\begin{tabular}{l l c c c c}
    	\hline
    	Category &Method &Ar$\xrightarrow{}$ Cl & Ar$\xrightarrow{}$ Pr & Ar$\xrightarrow{}$ Re &\textbf{Avg.}\\
    	\hline
    	\multirow{3}{*}{Single T.} 
    	&Baseline Network + Strong            &60.7 &77.7 &81.5 &73.3\\
    	&Baseline Network + Weak              &60.2 &78.3 &81.7 &73.4\\
    	&Baseline Network + Strong + Weak     &60.7 &78.2 &81.9 &73.6\\
    	\hline
    	\multirow{3}{*}{Multi T.} 
    	&Baseline Network + Strong            &60.7 &77.7 &81.3 &73.2\\
    	&Baseline Network + Weak              &60.4 &78.3 &82.0 &73.6\\
    	&Baseline Network + Strong + Weak     &61.2 &79.2 &81.8 &74.1\\
    	\hline
        \end{tabular}
    \end{table}

    \begin{figure*}
    	\centering
    	\footnotesize
    	\setlength{\tabcolsep}{0.1mm}
    	\begin{tabular}{cccc}
    		\includegraphics[width=0.24\linewidth]{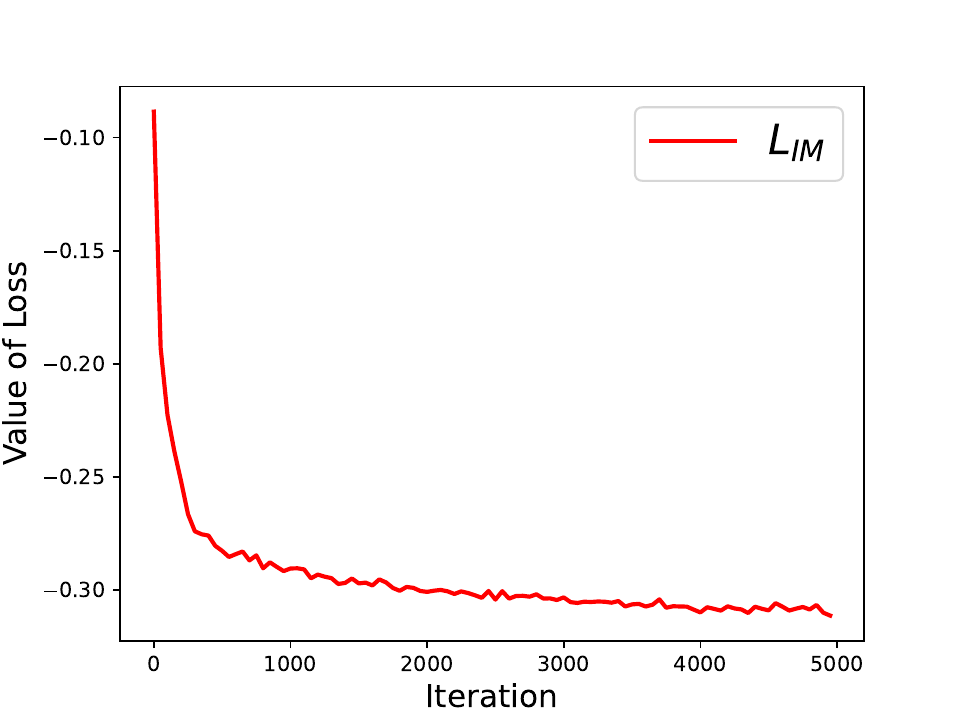}
    		& \includegraphics[width=0.24\linewidth]{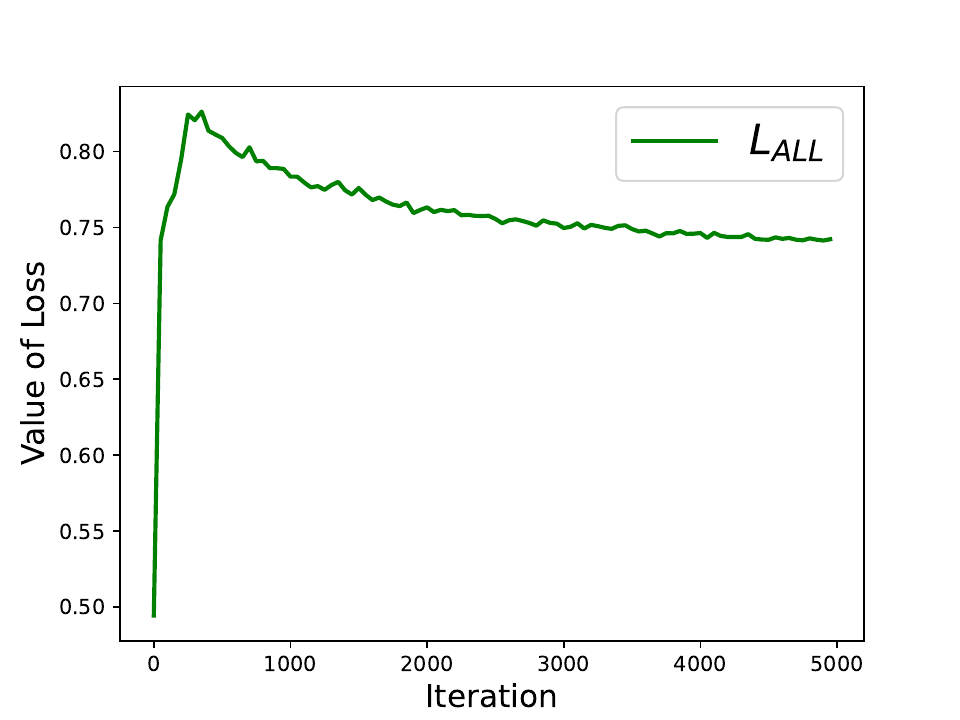}
    		& \includegraphics[width=0.24\linewidth]{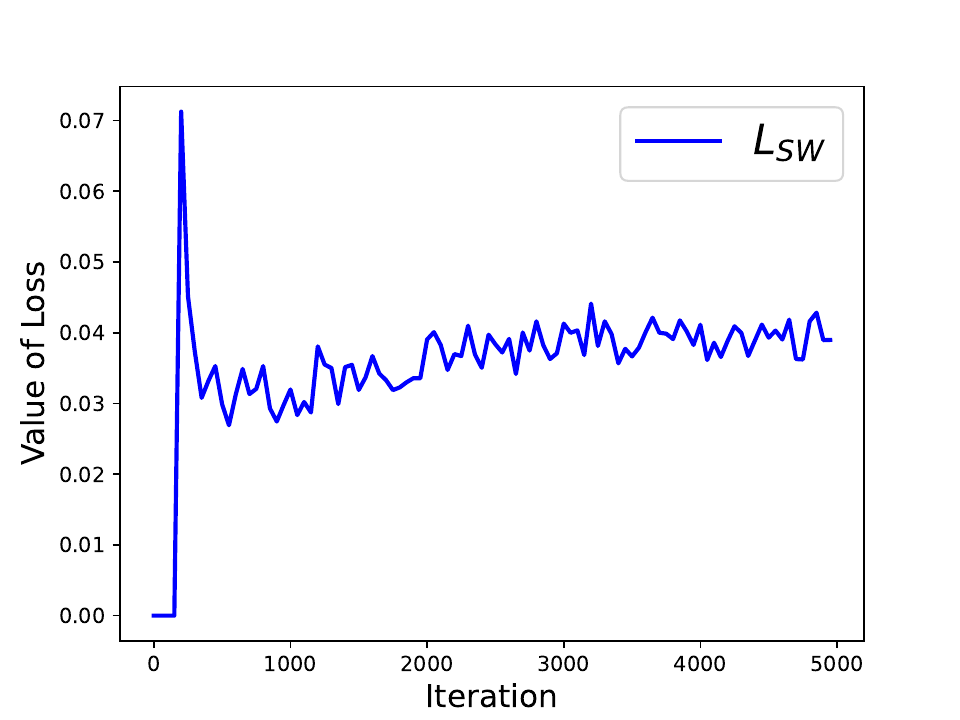}
    		& \includegraphics[width=0.24\linewidth]{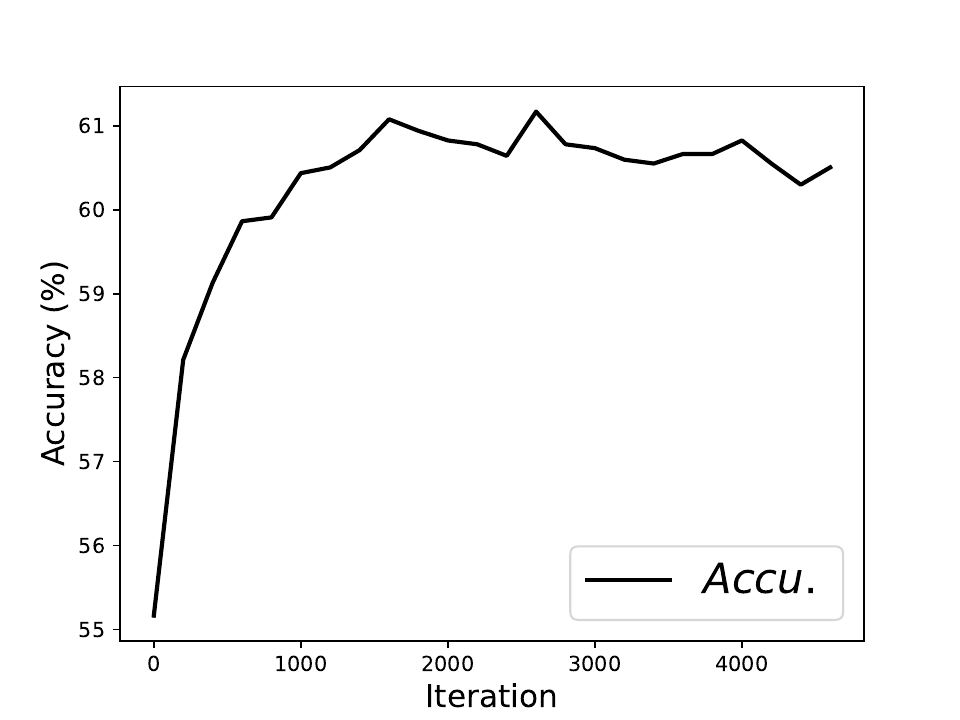}\\
    		(a) values of $L_{IM}$ &(b) values of $L_{ALL}$ &(c) values of $L_{SW}$ &(d) accuracy (\%) 
    	\end{tabular} 
    	\caption{Values of three loss terms $L_{IM}$, $L_{ALL}$, $L_{SW}$ and the accuracy during training procedure for the sub-task Ar$\xrightarrow{}$ Cl in Office-Home dataset.}
    	\label{Fig:curve_loss}
    \end{figure*} 

    \begin{figure}
    	\centering
    	\footnotesize
    	\begin{tabular}{c}
    		\includegraphics[width=0.95\linewidth]{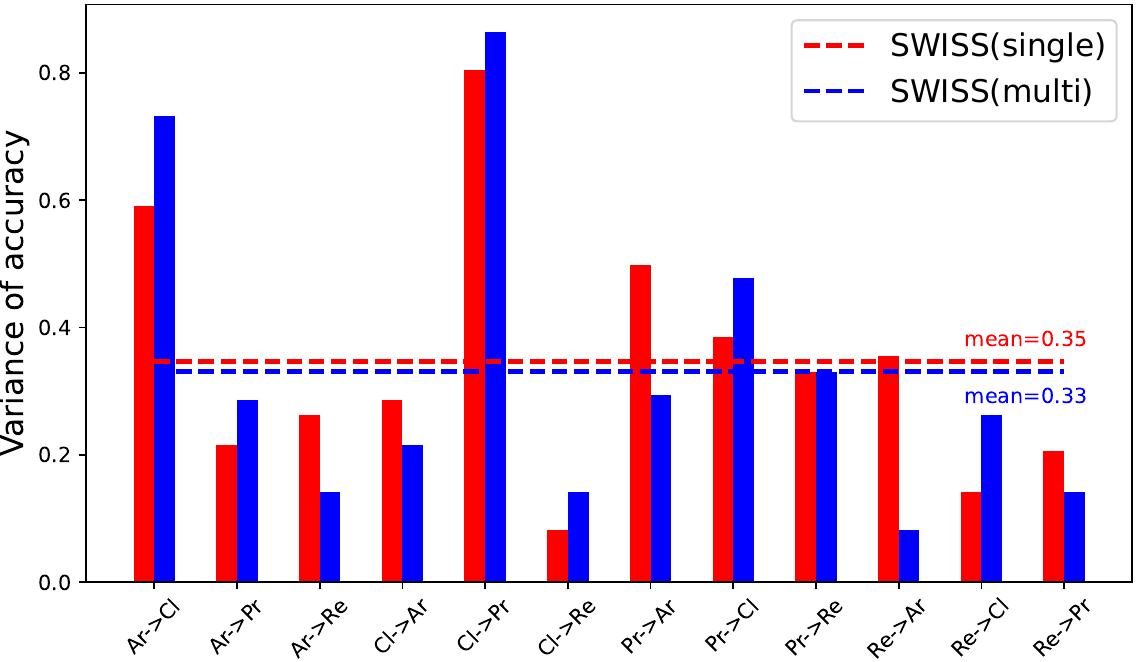}
    	\end{tabular} 
    	\caption{Variances of accuracy for each sub-task in Office-Home for both single-target and multi-target setting.}
    	\label{Fig:accu_variance}
    \end{figure}

    \begin{figure*}
    	\centering
    	\footnotesize
    	\setlength{\tabcolsep}{0.1mm}
    	\begin{tabular}{ccc}
    		\includegraphics[width=0.3\linewidth]{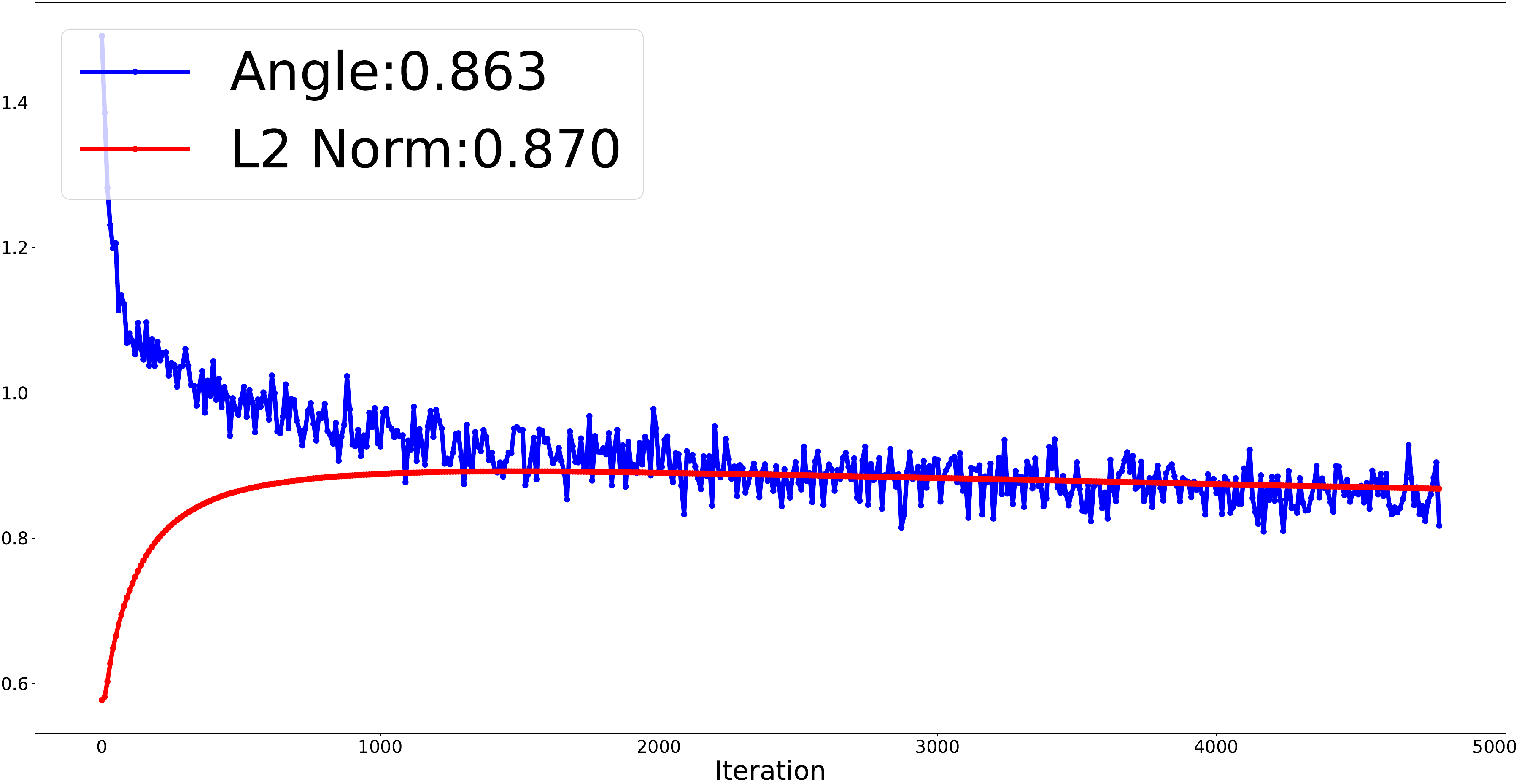}
    		& \includegraphics[width=0.3\linewidth]{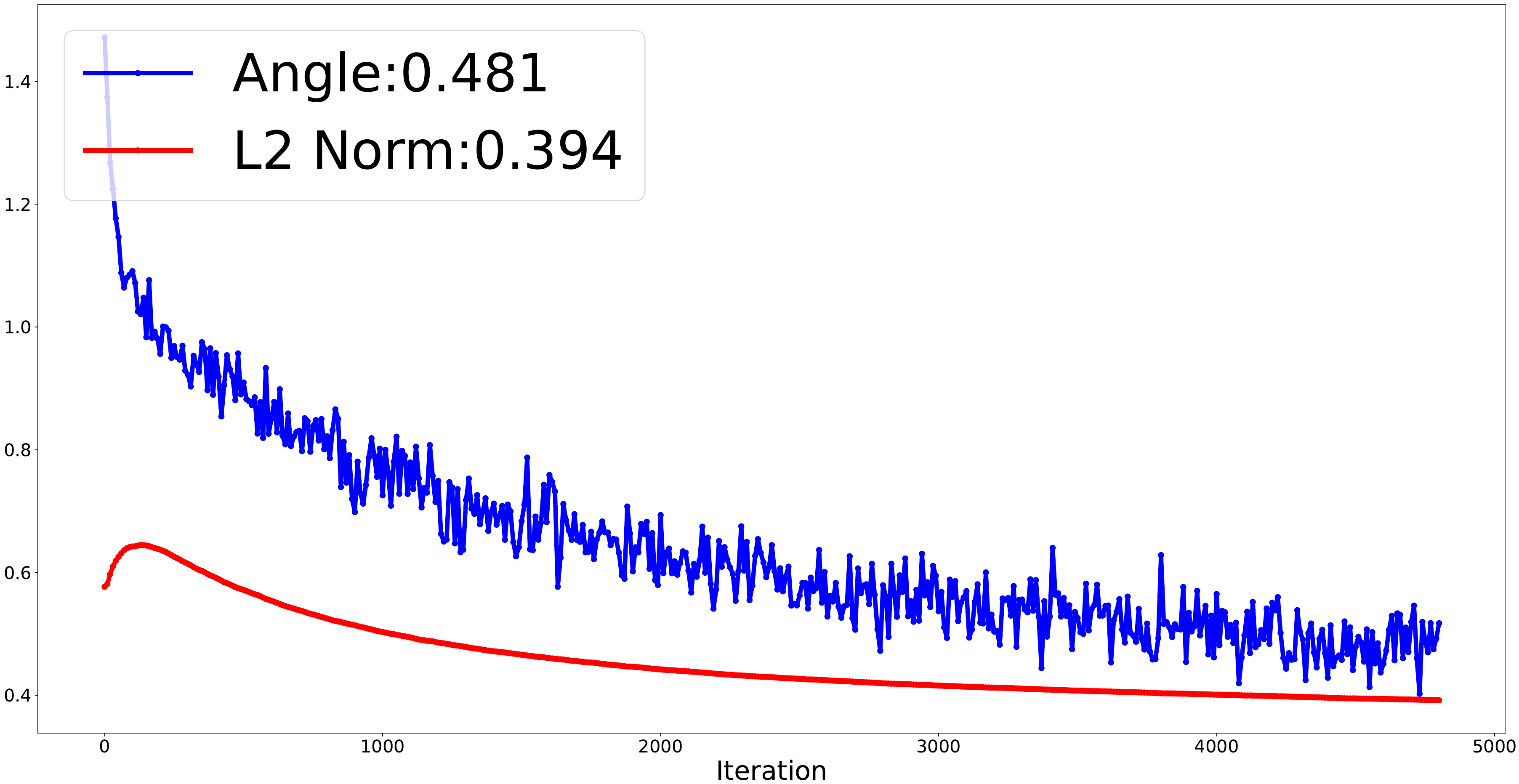}
    		& \includegraphics[width=0.3\linewidth]{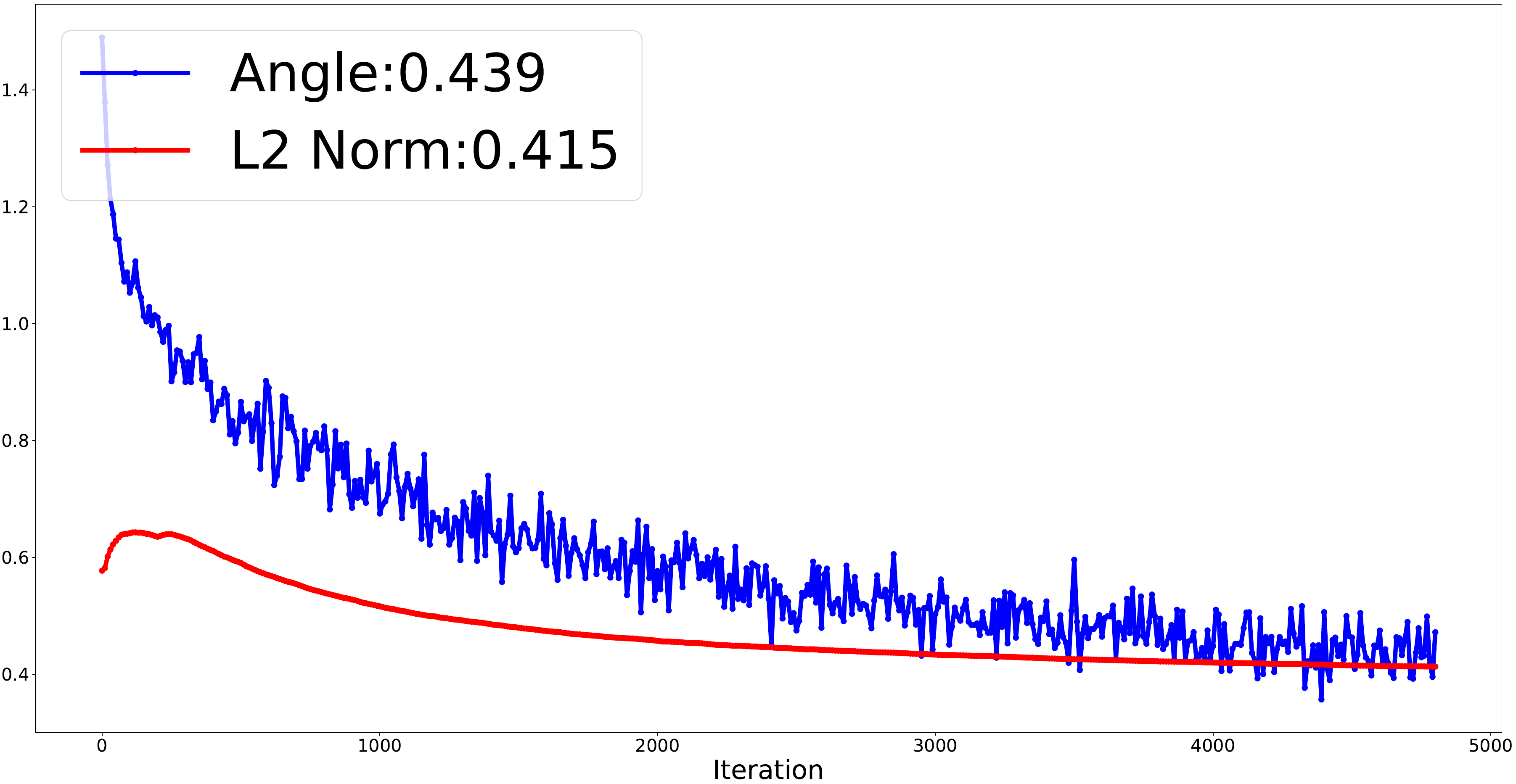} \\
    		\includegraphics[width=0.3\linewidth]{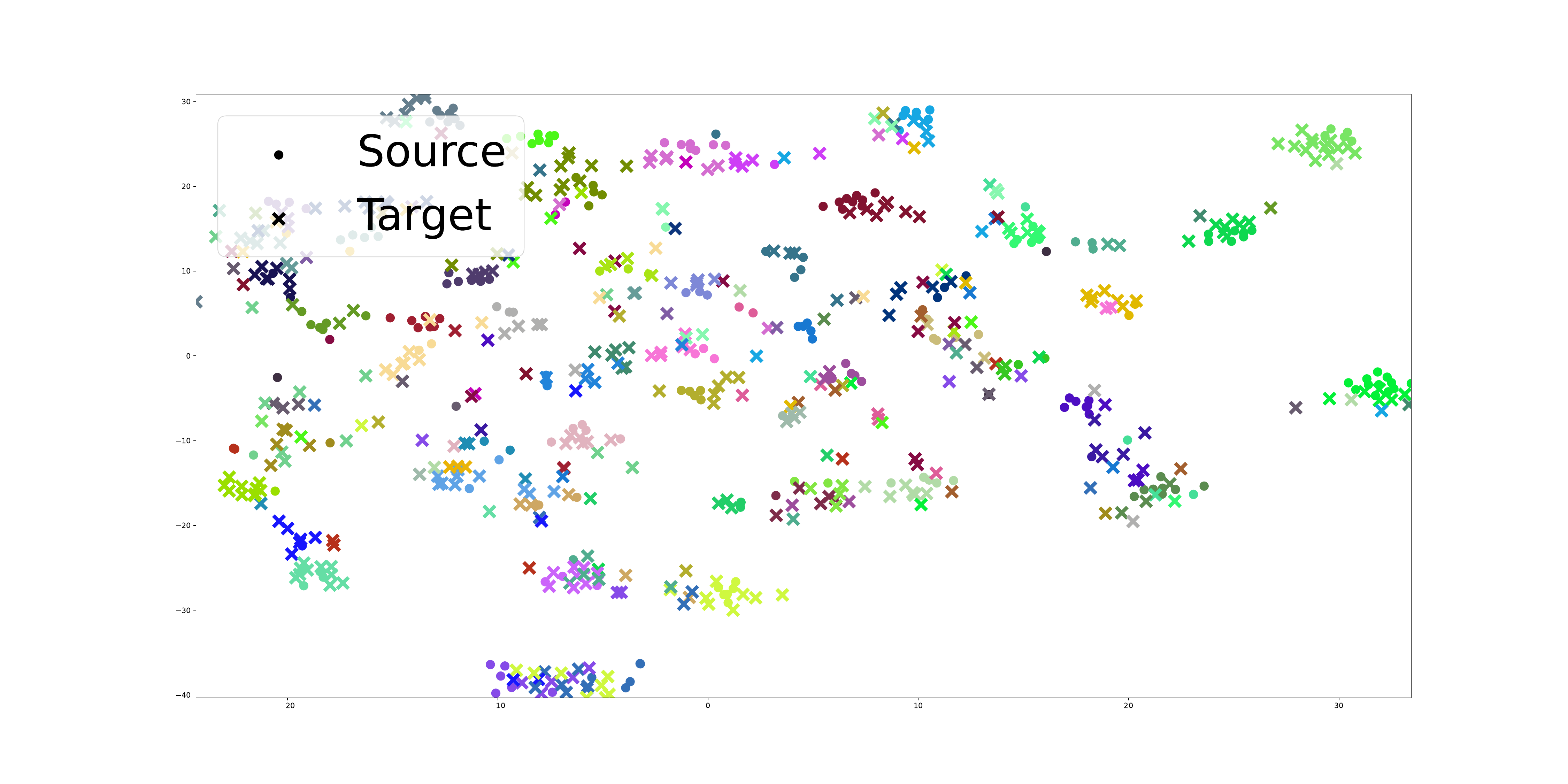}
    		& \includegraphics[width=0.3\linewidth]{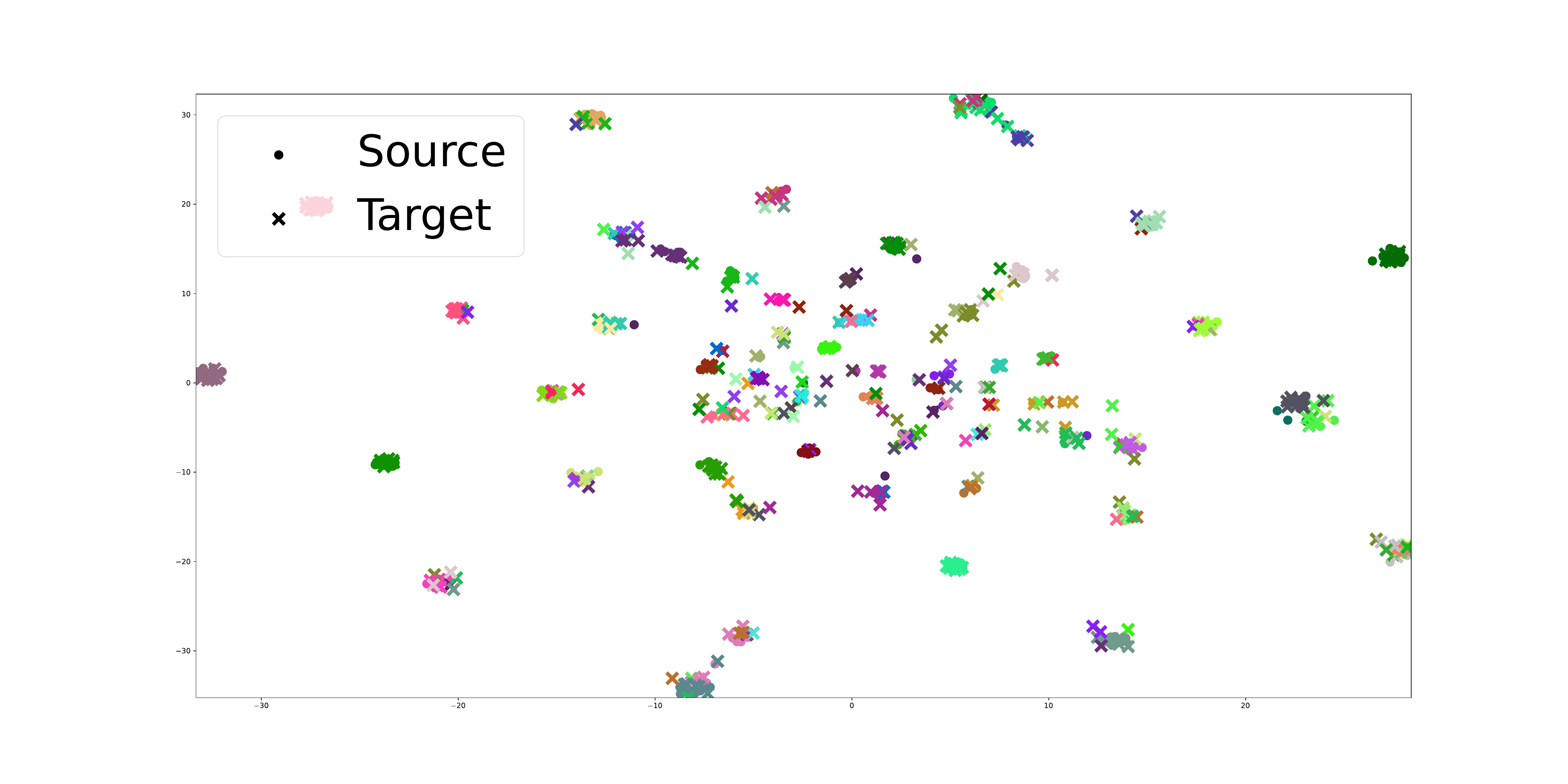}
    		& \includegraphics[width=0.3\linewidth]{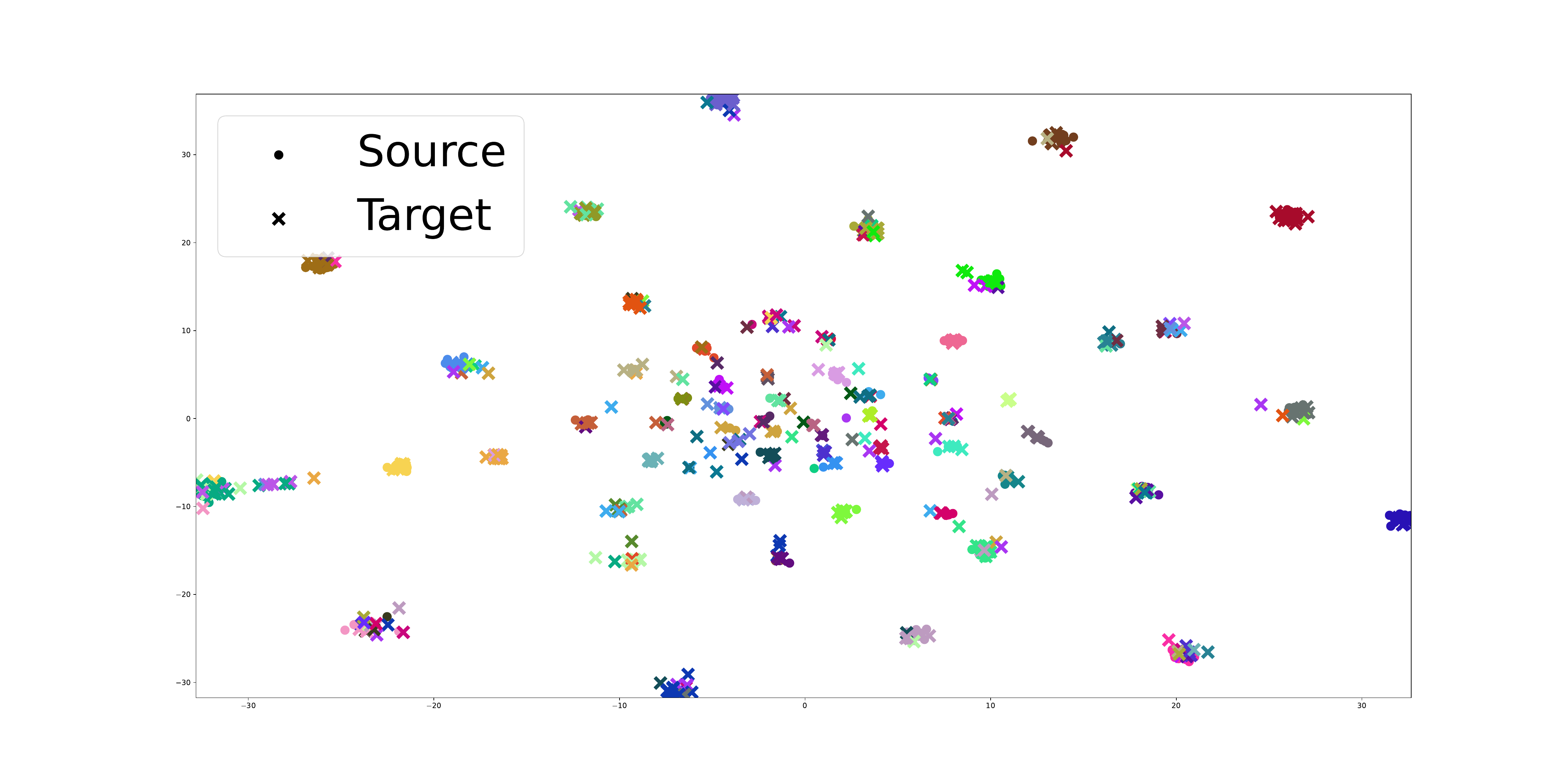} \\
    		(a) $L_{IM}$ &(b) $L_{IM}+L_{ALL}$ &(c) $L_{IM}+L_{ALL}+L_{SW}$
    	\end{tabular} 
    	\caption{Values of feature-prototype angle $\theta$ and $L_2$ norm of the prototype in \equ~\eqref{equ:logit}, and the distribution of feature vectors in source and target domain under different configurations of loss terms for the sub-task Ar$\xrightarrow{}$ Cl in Office-Home dataset. For the distributions in the second rows, only 10\% samples in the source and target domains are utilized for better view, we also use different colors to denote different classes.}
    	\label{Fig:tsne_curve}
    \end{figure*}

\textbf{Ablation study on different losses.}
\tab~\ref{tab:ablation_loss} shows the classification accuracies on the Office-Home dataset with different configurations of loss functions. Comparing the results to the baseline ResNet50~\cite{he2016deep}, we can observe the effectiveness of the mutual information maximization loss $L_{IM}$, which is adopted from previous works~\cite{li2021transferable,liang2020we}. Furthermore, we can see clear improvements when incorporating the proposed adversarial logit loss $L_{ALL}$ and the strong-weak integrated semi-supervision loss $L_{SW}$. By comparing the configurations $+L_{IM}+L_{ALL}$ and $+L_{IM}+L_{SW}$ with $+L_{IM}$, we achieve improvements of 1.2\% and 1.1\% for $L_{ALL}$ and $L_{SW}$, respectively. Moreover, the configuration $+L_{ALL}+L_{SW}$ achieves similar performance to $+L_{IM}+L_{ALL}$ and $+L_{IM}+L_{SW}$, which highlights the effectiveness of the proposed loss functions. By combining all three loss functions, we achieve the best performance among all the configurations, with an average improvement of 2.2\% over $L_{IM}$. This significant improvement demonstrates the advantages of our proposed method.

\textbf{Ablation study on network components.}
To evaluate the effectiveness of the proposed strong-weak integrated semi-supervision strategy, we conducted experiments on the Office-Home dataset with Ar as the source domain. \tab~\ref{tab:ablation_component} shows the accuracies of different configurations of network components, namely, strong supervision only, weak supervision only, and strong-weak integrated semi-supervision. Comparing ``Baseline Network + Strong" with ``Baseline Network + Weak", we observe that strong supervision tends to outperform weak supervision in the challenging task Ar$\xrightarrow{}$ Cl but underperforms in the easier tasks Ar$\xrightarrow{}$ Pr and Ar$\xrightarrow{}$ Re. This can be rationalized by the fact that strong supervision is more reliable than weak supervision when the prediction error is relatively high in challenging cases. However, in the easy cases, weak supervision provides similar prediction accuracy but offers greater sample variety. As a result, weak supervision can provide more informative signals, leading to better performance. The combination of strong supervision and weak supervision achieves the best performance in both the single-target and multi-target settings, with a margin of 0.35\% in average. This result demonstrates the effectiveness of the proposed strong-weak integrated semi-supervision strategy.

\textbf{Training stability.}
We investigated the accuracy and the values of three loss terms, namely $L_{IM}$, $L_{ALL}$, and $L_{SW}$, during the training process for the Ar$\xrightarrow{}$ Cl task. Fig.~\ref{Fig:curve_loss} illustrates the trends of these loss terms over iterations. Regarding $L_{IM}$, we observed a rapid decrease in its value, which converged after approximately 1000 iterations. This behavior is consistent with the findings in the work of~\cite{liang2020we}. For the proposed adversarial logit loss $L_{ALL}$, we observed that it initially experienced a significant jump in value within the first 500 iterations. This jump is caused by the optimization of the source domain loss term in \equ~\eqref{equ:l_ce}, which increases the logit values to generate high probabilities for the source domain samples. Subsequently, the proposed adversarial logit loss continuously reduces the logit values, resulting in a gradual decrease and convergence of the loss after approximately 3000 iterations. As for $L_{SW}$ shown in Fig.~\ref{Fig:curve_loss}(c), its value remained at 0 for the first 200 iterations, indicating that it was not activated initially as mentioned in Sec.~\ref{sec:strong_weak_fusion}. Afterward, we observed a rapid drop in the value of $L_{SW}$ from a large initial value, followed by convergence to a very small value. Regarding the testing accuracy shown in Fig.~\ref{Fig:curve_loss}(d), we observed a quick increase in accuracy during the initial training iterations, which then converged to a value between 60.0\% and 61.0\%. Additionally, Fig.~\ref{Fig:accu_variance} displays the variances of the testing accuracy. We observed that, except for the Ar$\xrightarrow{}$ Cl and Cl$\xrightarrow{}$ Pr sub-tasks, the variances were generally smaller than the average values. Specifically, the average variances were 0.35 and 0.33 for the single-target and multi-target settings, respectively.
    
\textbf{Visualization.}
To visualize the distributions of feature representations after domain adaptation, we employed the t-SNE method~\cite{maaten2008visualizing}. Fig.\ref{Fig:tsne_curve} illustrates the values of the angle $\theta$ and $L_2$ norm in \equ~\eqref{equ:logit}, as well as the feature distribution in the source and target domains under different configurations of loss terms. From the first row of Fig.\ref{Fig:tsne_curve}, we observed that the proposed adversarial loss $L_{ALL}$ effectively reduced the $L_2$ norm of the prototype and the feature-prototype angle, as indicated by the decreasing values of $\theta$ and the $L_2$ norm. This reduction suggests a smaller intra-class domain divergence. Moreover, by comparing the third column with the second column, we observed that the addition of $L_{SW}$ further reduced the angle. These findings are consistent with the results obtained from the t-SNE figures shown in the second row of Fig.\ref{Fig:tsne_curve}. Analyzing the t-SNE figures, we observed that the feature distribution obtained using only $L_{IM}$ appeared relatively sparse. However, incorporating $L_{ALL}$ made the distribution more compact within each class and more distinct/separated among different classes. Furthermore, the addition of $L_{SW}$ led to further concentration of the feature distribution. Overall, these results demonstrate the effectiveness of the proposed loss terms in reducing intra-class domain divergence and enhancing the compactness and separability of the feature representations.

\section{Conclusion}
This paper presents an unified framework which adopts a novel strong-weak integrated semi-supervision strategy for both single target and multi target domain adaptation. A strong representative set with high prediction confidence but low diversity and a weak representative set with low prediction confidence but high diversity are maintained and updated during the training process. The fusion of them generates augmented training samples with pseudo-label, which are utilized as supervision for training the network. The extension from single to multi target domain adaptation is achieved by adopting the peer scaffolding strategy, which updates the strong representative set with samples not only from the target domain itself but also from its peer domains. Moreover, a novel adversarially optimized loss based on the logit instead of the probability is developed to further reduce the intra-class domain divergence. Comprehensive experiments on several popular benchmarks have demonstrated the effectiveness of the proposed method. In the future, the extension from our work to multi source domain adaptation can also be investigated.

{
\bibliographystyle{IEEEtran}
\bibliography{egbib}
}




\vfill

\end{document}